\documentclass[lettersize,journal]{IEEEtran}

\usepackage{epsfig}
\usepackage{graphicx}
\usepackage{subfigure}
\usepackage{amssymb}
\usepackage{amsmath}
\usepackage{array}
\usepackage{booktabs}
\usepackage{bbm}
\usepackage{url}
\usepackage[T1]{fontenc}
\usepackage{multirow}
\usepackage{bbding}
\usepackage{indentfirst}
\usepackage{algorithmicx,algorithm}
\usepackage{makecell}
\usepackage[noend]{algpseudocode}

\newcommand{\PreserveBackslash}[1]{\let\temp=\\#1\let\\=\temp}
\newcolumntype{C}[1]{>{\PreserveBackslash\centering}p{#1}}
\newcolumntype{R}[1]{>{\PreserveBackslash\raggedleft}p{#1}}
\newcolumntype{L}[1]{>{\PreserveBackslash\raggedright}p{#1}}

\usepackage{tikz,xcolor}
\usepackage[pagebackref=true,breaklinks=true,letterpaper=true,colorlinks,bookmarks=false]{hyperref}

\definecolor{lime}{HTML}{A6CE39}

\DeclareRobustCommand{\orcidicon}{%
\begin{tikzpicture}
\draw[lime, fill=lime] (0,0) 
circle [radius=0.16] 
node[white] {{\fontfamily{qag}\selectfont \tiny $\dot{\mathsf I}$D}};
\draw[white, fill=white] (-0.0625,0.095) 
circle [radius=0.007];
\end{tikzpicture}
\hspace{-2mm}
}

\foreach \x in {A, ..., Z}{%
\expandafter\xdef\csname orcid\x\endcsname{\noexpand\href{https://orcid.org/\csname orcidauthor\x\endcsname}{\noexpand\orcidicon}}
}

\usepackage{ifpdf}
\usepackage{cite}
\usepackage{amsmath}
\usepackage{fixltx2e}
\usepackage{stfloats}

\ifCLASSOPTIONcaptionsoff
\usepackage[nomarkers]{endfloat}
\let\MYoriglatexcaption\caption
\renewcommand{\caption}[2][\relax]{\MYoriglatexcaption[#2]{#2}}
\fi

\usepackage{url}

\begin{document}
\title{SegGroup: Seg-Level Supervision for 3D Instance and Semantic Segmentation}

\author{An Tao\orcidA{},~\IEEEmembership{Graduate Student Member,~IEEE,}
	Yueqi Duan\orcidB{},~\IEEEmembership{Member,~IEEE,}
	Yi Wei\orcidC{},~\\
	Jiwen Lu\orcidD{},~\IEEEmembership{Senior Member,~IEEE,}
	and~Jie~Zhou\orcidE{},~\IEEEmembership{Senior Member,~IEEE}% <-this % stops a space
	
	\thanks{Manuscript received 9 July 2021; revised 22 March 2022 and 4 June 2022; accepted 1 July 2022. This work was supported in part by the National Key Research and Development Program of China under Grant 2017YFA0700802, in part by the National Natural Science Foundation of China under Grant 62125603 and Grant U1813218, and in part by the Beijing Academy of Artificial Intelligence (BAAI). The associate editor coordinating the review of this manuscript and approving it for publication was Prof. Adrian Munteanu.}
	\thanks{\textit{(Corresponding author: Yueqi Duan.)}}
	\thanks{An Tao, Yi Wei, Jiwen Lu, and Jie Zhou are with the Department of Automation, Tsinghua University, Beijing 100084, China, and also with the Beijing National Research Center for Information Science and Technology (BNRist), Beijing 100084, China (e-mail: ta19@mails.tsinghua.edu.cn; y-wei19@mails.tsinghua.edu.cn; lujiwen@tsinghua.edu.cn; jzhou@tsinghua.edu.cn).}
	\thanks{Yueqi Duan is with the Department of Electronic Engineering, Tsinghua University, Beijing 100084, China (e-mail: duanyueqi@tsinghua.edu.cn).}
	\thanks{Digital Object Identifier 10.1109/TIP.2022.3190709}}% <-this % stops a space
	%\thanks{Manuscript received April 19, 2005; revised August 26, 2015.}}% <-this % stops a space

\markboth{IEEE TRANSACTIONS ON IMAGE PROCESSING}%
{Tao \MakeLowercase{\textit{et al.}}: SegGroup: Seg-Level Supervision for 3D Instance and Semantic Segmentation}

\maketitle

\begin{abstract}
	Most existing point cloud instance and semantic segmentation methods rely heavily on strong supervision signals, which require point-level labels for every point in the scene. However, such strong supervision suffers from large annotation costs, arousing the need to study efficient annotating. In this paper, we discover that the locations of instances matter for both instance and semantic 3D scene segmentation. By fully taking advantage of locations, we design a weakly-supervised point cloud segmentation method that only requires clicking on one point per instance to indicate its location for annotation. With over-segmentation for pre-processing, we extend these location annotations into segments as seg-level labels. We further design a segment grouping network (SegGroup) to generate point-level pseudo labels under seg-level labels by hierarchically grouping the unlabeled segments into the relevant nearby labeled segments, so that existing point-level supervised segmentation models can directly consume these pseudo labels for training. Experimental results show that our seg-level supervised method (SegGroup) achieves comparable results with the fully annotated point-level supervised methods. Moreover, it outperforms the recent weakly-supervised methods given a fixed annotation budget. Code is available at \url{https://github.com/antao97/SegGroup}.
\end{abstract}

\begin{IEEEkeywords}
	Point cloud segmentation, seg-level supervision, weakly-supervised learning, graph neural network
\end{IEEEkeywords}

% \fi
%\IEEEpeerreviewmaketitle

\section{Introduction}

\IEEEPARstart{R}{ecent} years have witnessed significant progress on analyzing different 3D geometric data structures, including point cloud~\cite{qi2017pointnet,qi2017pointnet++}, mesh~\cite{hanocka2019meshcnn,schult2020dualconvmesh}, voxel grid~\cite{qi2016volumetric,wu20153d}, multi-view~\cite{dai20183dmv,su2015multi} and implicit function~\cite{park2019deepsdf,mescheder2019occupancy,duan2020curriculum}. Due to the popularity of varying scanning devices, 3D point cloud data is easy to obtain and thus arouses increasingly attention. Recently, many deep learning methods have been proposed to directly operate on point clouds and have achieved encouraging performance~\cite{qi2017pointnet,qi2017pointnet++,wu2019pointconv,wang2019dynamic,thomas2019kpconv,cheng2021net}.

Point cloud instance and semantic segmentation are two fundamental but challenging tasks in 3D scene understanding. Given a point cloud, instance segmentation aims to find all existing objects and mark each object with a unique instance label and a semantic class, while semantic segmentation only predicts semantic classes. Over the past few years, strong point-level supervisions that annotate every point in the scene have derived rapid performance improvement on point cloud instance and semantic segmentation tasks~\cite{wang2018sgpn,yi2019gspn,hou20193d,yang2019learning,jiang2020pointgroup,choy20194d}. However, as each scene may contain a large number of points, it is highly time-consuming to annotate all the points. For example, ScanNet~\cite{dai2017scannet} is a widely used large-scale real-world indoor dataset. It contains 1,613 scenes and each scene has 150,000 points on average. Even though ScanNet adopts over-segmentation to reduce the annotation workload, it still needs around 22.3 minutes to annotate all the segments in one scene. With a growing number of unlabeled 3D point cloud data in real-world applications, a natural question is raised: is it necessary to label all points in a point cloud scene?

\begin{figure}[t]
	\centering
	\vspace{-7pt}
	\subfigure[Segments~~~~~~]{
		\begin{minipage}[b]{0.42\linewidth}
			\includegraphics[width=1\linewidth, trim=0 0 0 0,clip]{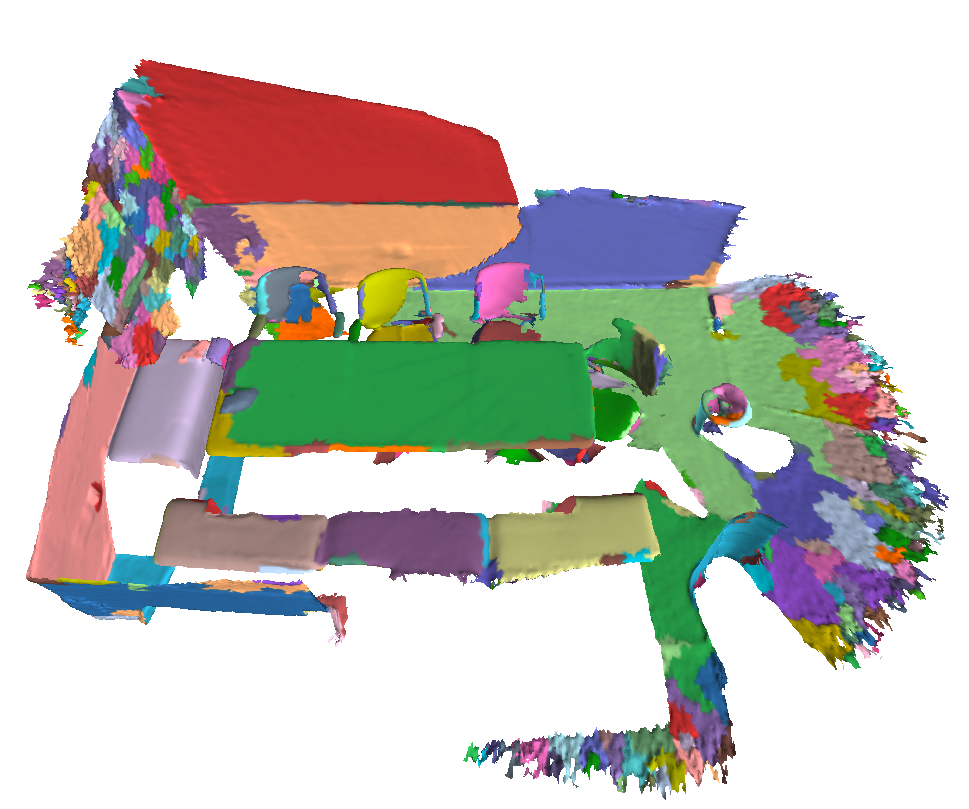}
		\end{minipage}
	}
	\subfigure[Location Annotations]{
		\begin{minipage}[b]{0.5\linewidth}
			\includegraphics[width=1\linewidth, trim=0 0 0 0,clip]{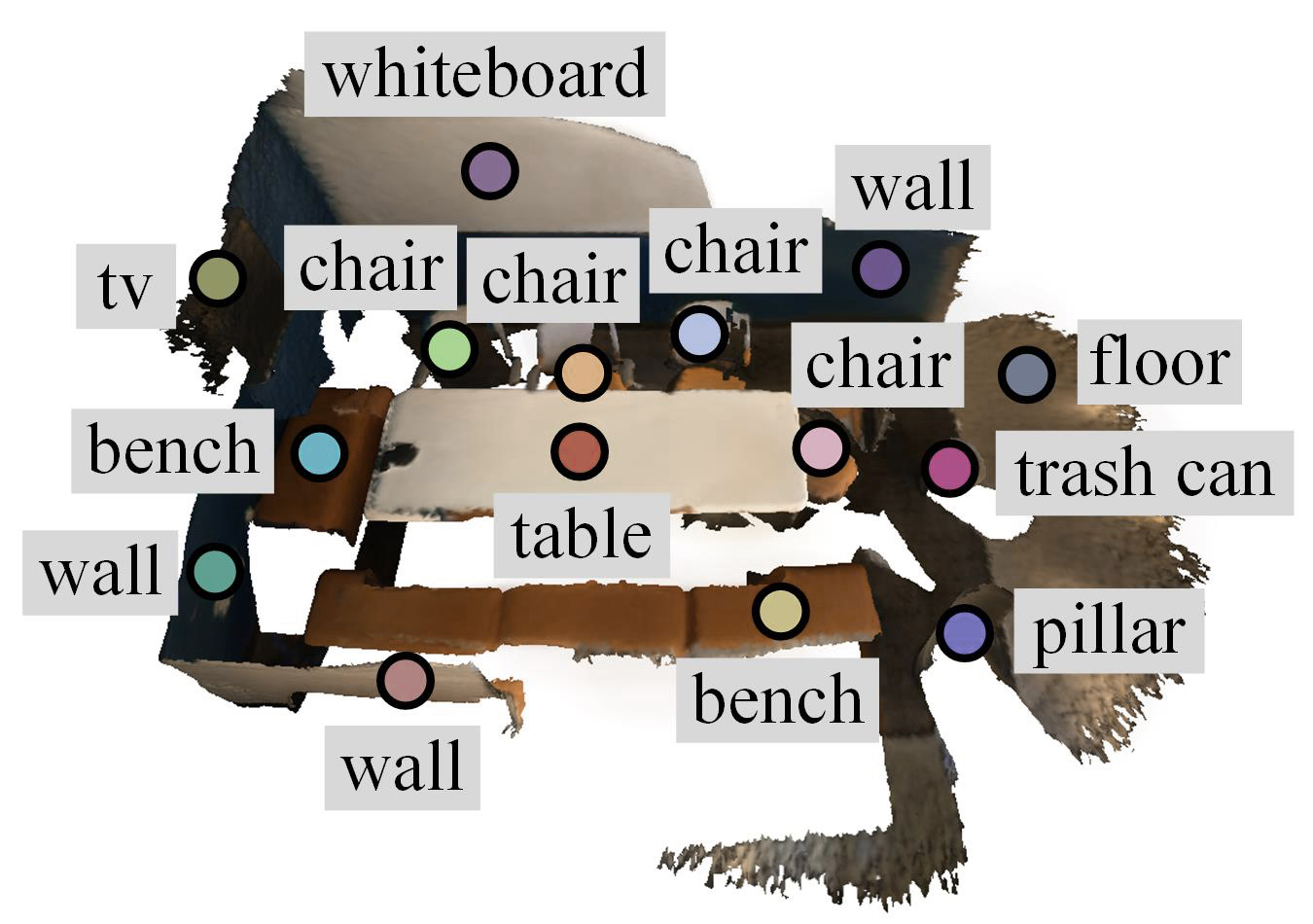}
			
		\end{minipage}
		\hspace{-10pt}
	}
	\subfigure[Seg-level Labels]{
		\hspace{1pt}
		\begin{minipage}[b]{0.42\linewidth}
			\includegraphics[width=1\linewidth, trim=0 0 0 20,clip]{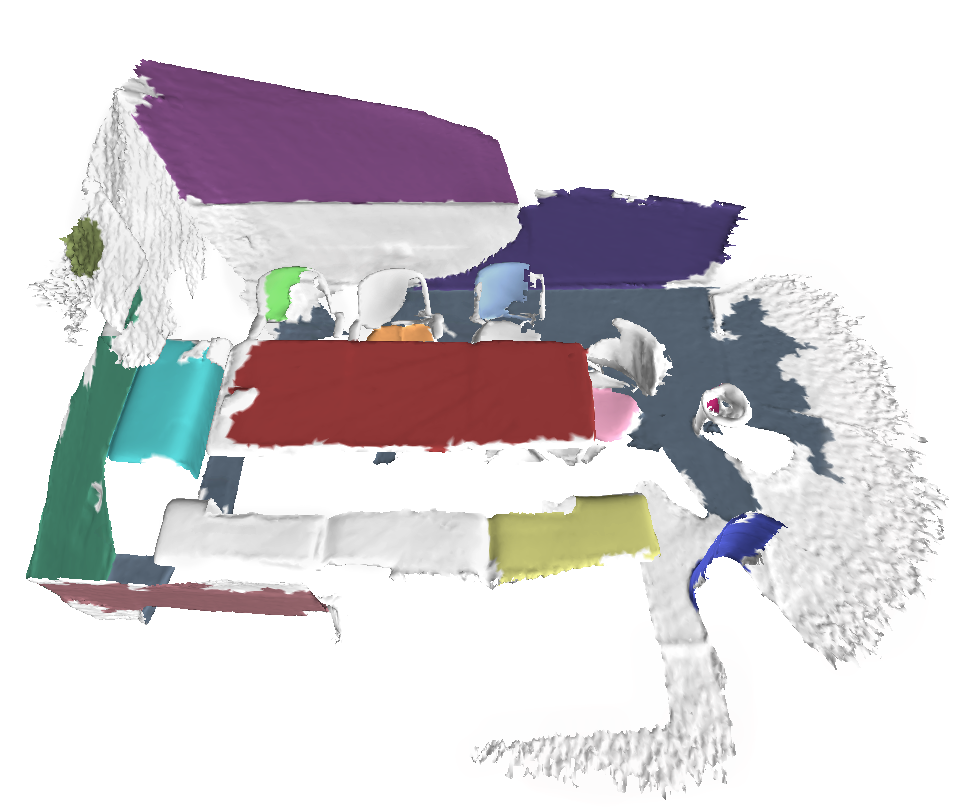}
		\end{minipage}
	}
	\subfigure[Point-level Labels]{
		\hspace{9pt}
		\begin{minipage}[b]{0.42\linewidth}
			\includegraphics[width=1\linewidth, trim=0 0 0 20,clip]{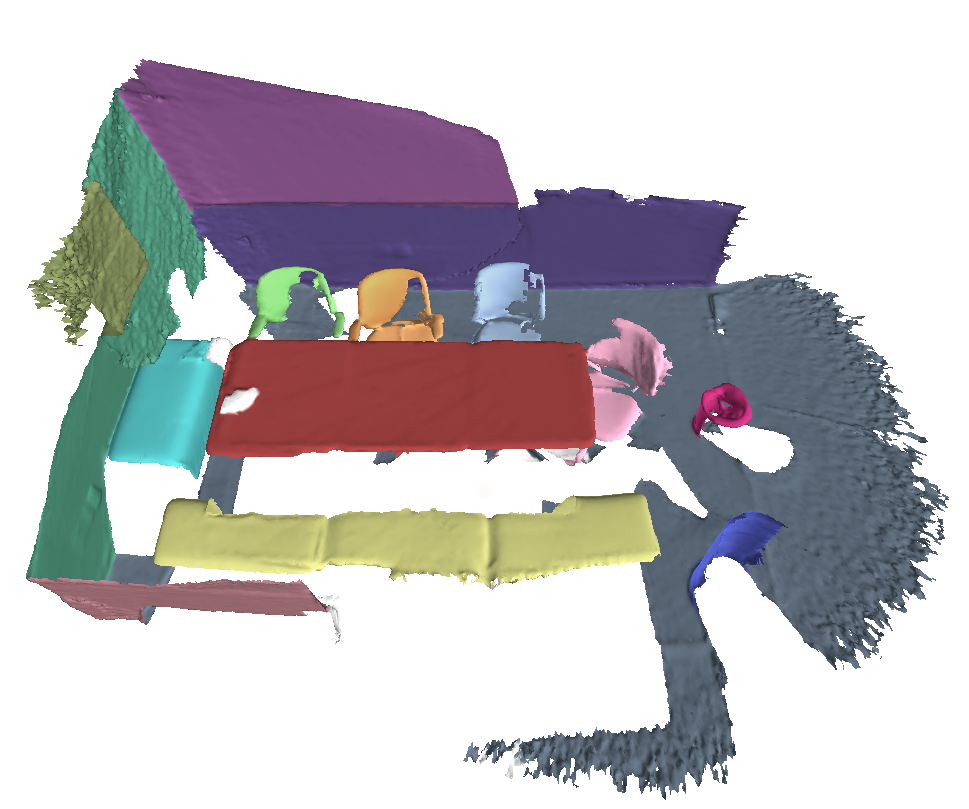}
		\end{minipage}
		
	}
	\vspace{7pt}
	\caption{Illustration of annotation process to give instance locations. (a) Given a scene, we first perform over-segmentation to get segments as the data pre-processing step. (b) We click one point on the most representative segment of each instance to give the location and annotate it with semantic classes and instance IDs. (c) Then, we extend location annotations into segments to obtain seg-level labels. (d) Compared with point-level labels which require 22.3 minutes per scene to annotate all points, seg-level labels only require 1.93 minutes per scene to give location information of each instance.
		%	Illustration of annotation process for seg-level label. (a) Given a scene, we first perform over-segmentation to get segments as data pre-processing step. (b) We click on one point per instance to annotate semantic classes and instance IDs. (c) Then, we map point annotations into segments to obtain seg-level labels. (d) Compared with point-level labels, seg-level labels only require 8.65\% annotation time to label 31.06\% points.
	}
	\label{seg_label}
	%	\vspace{-5pt}
\end{figure}

To alleviate the need for detailed point-level annotations, we focus on weak supervision in point cloud instance and semantic segmentation, which is far easier to obtain. Few prior works study weak supervision in point cloud segmentation~\cite{wei2020multi,xu2020weakly,hou2021exploring}, where scene-level label, subcloud-level label, and point annotation are three recent weak label forms. Scene-level labels~\cite{wei2020multi} indicate the semantic classes appearing in a point cloud scene, while subcloud-level labels~\cite{wei2020multi} indicate the semantic classes appearing in a spherical subcloud sampled from a point cloud. Point annotations~\cite{xu2020weakly,hou2021exploring, hu2021sqn, zhang2021perturbed} are a set of labeled points sampled from a point cloud randomly or by a specially designed algorithm. Although these weak label forms reduce annotation effort significantly, they cannot provide any instance-specific information, which makes them not suitable for point cloud instance segmentation. A recent weakly-supervised point cloud semantic segmentation method OTOC~\cite{liu2021one} uses point annotations per instance, but it cannot be directly applied to the instance segmentation task.
%Although very recently a weakly point cloud semantic segmentation method OTOC~\cite{liu2021one} adopts the point annotations that are annotated one point per instance, it cannot be directly applied to the instance segmentation task due to the semantic nature of its energy function.

%For weak label in point cloud segmentation, Wei \textit{et al.} proposed scene-level and subcloud-level label with reference to image-level label in 2D image. Scene-level label indicates the classes appearing in the scene. Considering some classes frequently appear in almost all scenes, e.g. wall, floor, chair and table, to avoid this problem Wei \textit{et al.} subsampled several spherical subclouds in each scene to ensure class appearence diversity and define subcloud-level label to indicate the classes appearing in the subcloud. Although these two weak label forms reduce labeling cost successfully, their benefits are not high enough.

In this paper, we discover that \textit{the locations of instances matter for both instance and semantic 3D scene segmentation.} Compared with the 2D images which lack depth dimension, locations of instances can be more precisely in 3D scenes.
To fully take advantage of locations, we only need to click on one point per instance to indicate its location for segmentation, rather than labeling every point in the scene. With these locations, we design a weakly-supervised point cloud segmentation task to explore the importance of locations. 

Fig.~\ref{seg_label} illustrates the annotation process to give instance locations. Following the labeling strategy of point-level labels in ScanNet~\cite{dai2017scannet}, we first perform over-segmentation to obtain segments as pre-processing. Unlike 2D images which usually face occlusions and lightning variances, 3D data structures are suitable for over-segmentation for their apparent boundaries between different simple geometry parts. Then, we click one point on the most representative\footnote{In this work, we consider the largest segment of the instance as the most representative segment to indicate the instance location.} segment of each instance to give location annotations and extend them into corresponding segments as seg-level labels. Finally, the seg-level labels are adopted as supervision signals for point cloud segmentation.
According to our manual annotations, seg-level labels only require 1.93 minutes per scene to click 0.028\% points compared with the strong point-level labels. However, due to the over-segmentation, these annotated segments contain 29.42\% points of the whole scene.

To learn point cloud instance and semantic segmentation under seg-level supervision, we design a two-stage approach. We first propose a segment grouping network (SegGroup) to generate point-level pseudo labels from seg-level labels for the remaining unannotated points on the training set. Then, we adopt an existing point-level supervised point cloud segmentation model for standard training. The two stages are trained separately, and the evaluation of the segmentation performance is conducted on the existing point-level supervised model. In the SegGroup network, we propagate label information by grouping unlabeled segments into the relevant nearby labeled segments and conduct the grouping operation hierarchically. 
%Before each grouping step, we design a Graph Convolution Network (GCN)~\cite{kipf2016semi} to update the representation of each segment by aggregating representations of its similar neighbors. 
After all grouping operations, all points in the point cloud scene are assigned with labels that are considered as point-level pseudo labels.
%With the generated pseudo labels, the existing point-level supervised point cloud segmentation methods can directly consume these labels for training. 

Experimental results show that our seg-level supervised method (SegGroup) achieves comparable results with the fully annotated point-level supervised methods. It also outperforms the recent weakly-supervised methods~\cite{hou2021exploring,wei2020multi} given a fixed annotation budget. These results validate that annotating location information is a low-cost but high-yield labeling manner for 3D scene segmentation.

Our key contributions are summarized as follows:

\begin{itemize}
	\item[1)] We discover that the locations of instances matter for 3D scene segmentation. To fully take advantage of locations, we design a weakly-supervised point cloud segmentation method that only requires clicking on one point per instance to indicate its location for annotation.
	
	\item[2)] We design a segment grouping network (SegGroup) to generate pseudo labels for the remaining unannotated points. Specifically, the network propagates label information by grouping unlabeled segments into the relevant nearby labeled segments hierarchically. Then, the pseudo labels are used for standard fully supervised training.
	
	\item[3)] Experimental results show that our seg-level supervised method (SegGroup) achieves comparable results with the fully annotated point-level supervised methods. Moreover, it also outperforms the recent weakly-supervised methods given a fixed annotation budget.
\end{itemize}

\section{Related Work}
In this section, we briefly review two related topics: 1)
point cloud segmentation, and 2) weakly-supervised image segmentation.

\subsection{Point Cloud Segmentation} Approaches on point cloud semantic segmentation can be mainly classified into two categories: voxel-based~\cite{graham20183d,choy20194d} and point-based~\cite{qi2017pointnet,qi2017pointnet++,wang2019dynamic,thomas2019kpconv,shuai2021backward,li2020multi,liu2020semantic}. Voxel-based approaches voxelized point clouds into 3D grids in order to apply powerful 3D CNNs, while point-based approaches directly design models on point clouds to learn per-point local features. Recent methods on point-based scheme include neighbouring feature pooling~\cite{zhang2019shellnet,hu2020randla}, graph message passing~\cite{landrieu2018large}, attention-based aggregation~\cite{xie2018attentional,yang2019modeling}, and kernel-based convolution~\cite{thomas2019kpconv,wu2019pointconv}.
For point cloud instance segmentation, there are two common strategies to find instances in 3D scenes: detection-based~\cite{yi2019gspn,hou20193d} and segmentation-based~\cite{wang2018sgpn,jiang2020pointgroup,han2020occuseg}. Detection-based approaches first extract 3D bounding boxes using object detection techniques, and then find the object mask inside each box. By contrast, segmentation-based approaches first predict semantic labels for each point with a semantic segmentation framework, and then group points into different objects. 

While most existing methods heavily rely on strong point-level labels, only a few works have studied weakly-supervised point cloud semantic segmentation. Wei~\textit{et al.}~\cite{wei2020multi} proposed scene-level labels and subcloud-level labels to indicate the semantic classes appearing in a point cloud scene and a spherical subcloud sampled from a given point cloud, respectively.
Hou~\textit{et al.}~\cite{hou2021exploring} proposed a 3D pre-training method that makes use of both point-level correspondences and spatial contexts in a scene with annotations of a given number of points per scene by an active selection process with the pre-trained model.
Xu~\textit{et al.}~\cite{xu2020weakly}, Hu~\textit{et al.}~\cite{hu2021sqn}, and Zhang~\textit{et al.}~\cite{zhang2021perturbed} studied point cloud segmentation under a fraction of randomly labeled points.
Although these weak label forms reduce annotation effort significantly, they cannot provide any instance-specific location information, which is a very important supervision signal in the 3D scene. In contrast, our seg-level labels are annotated per instance and are easy to be adopted by annotators when they face the need for labeling.
Liu~\textit{et al.}~\cite{liu2021one} proposed a weakly-supervised point cloud semantic segmentation method OTOC by annotating one point per instance in the point cloud scene. In contrast, our method can be generally applied to semantic and instance segmentation tasks.

\begin{figure*}[t]
	\centering
	\subfigure[Annotating from Scratch~~~~~~]{
		\label{anno_a}
		\begin{minipage}[b]{0.315\linewidth}
			\includegraphics[width=1\linewidth]{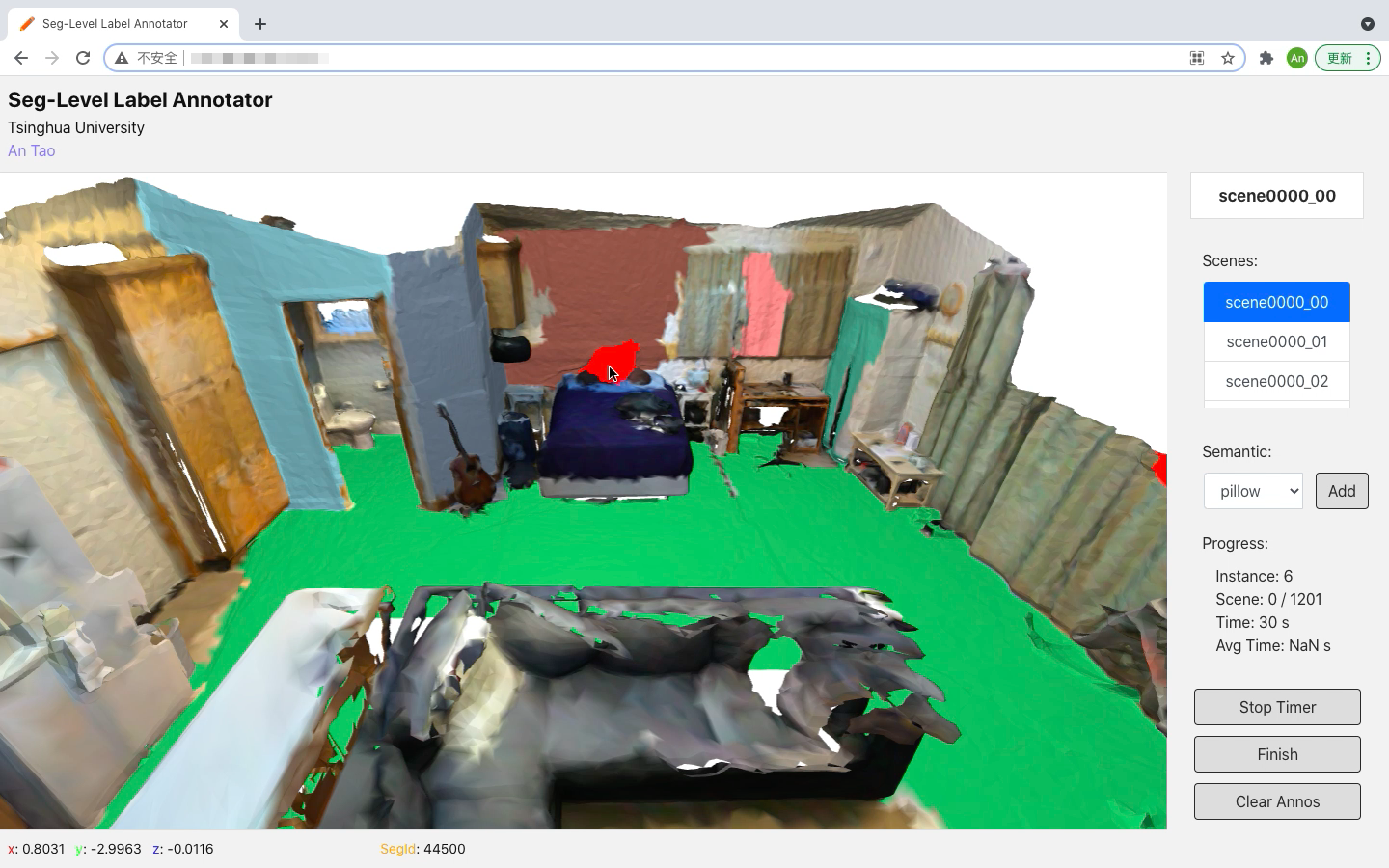}
		\end{minipage}
	}
	\subfigure[Annotating with GT Labels~~~~~~]{
		\label{anno_b}
		\begin{minipage}[b]{0.315\linewidth}
			\includegraphics[width=1\linewidth]{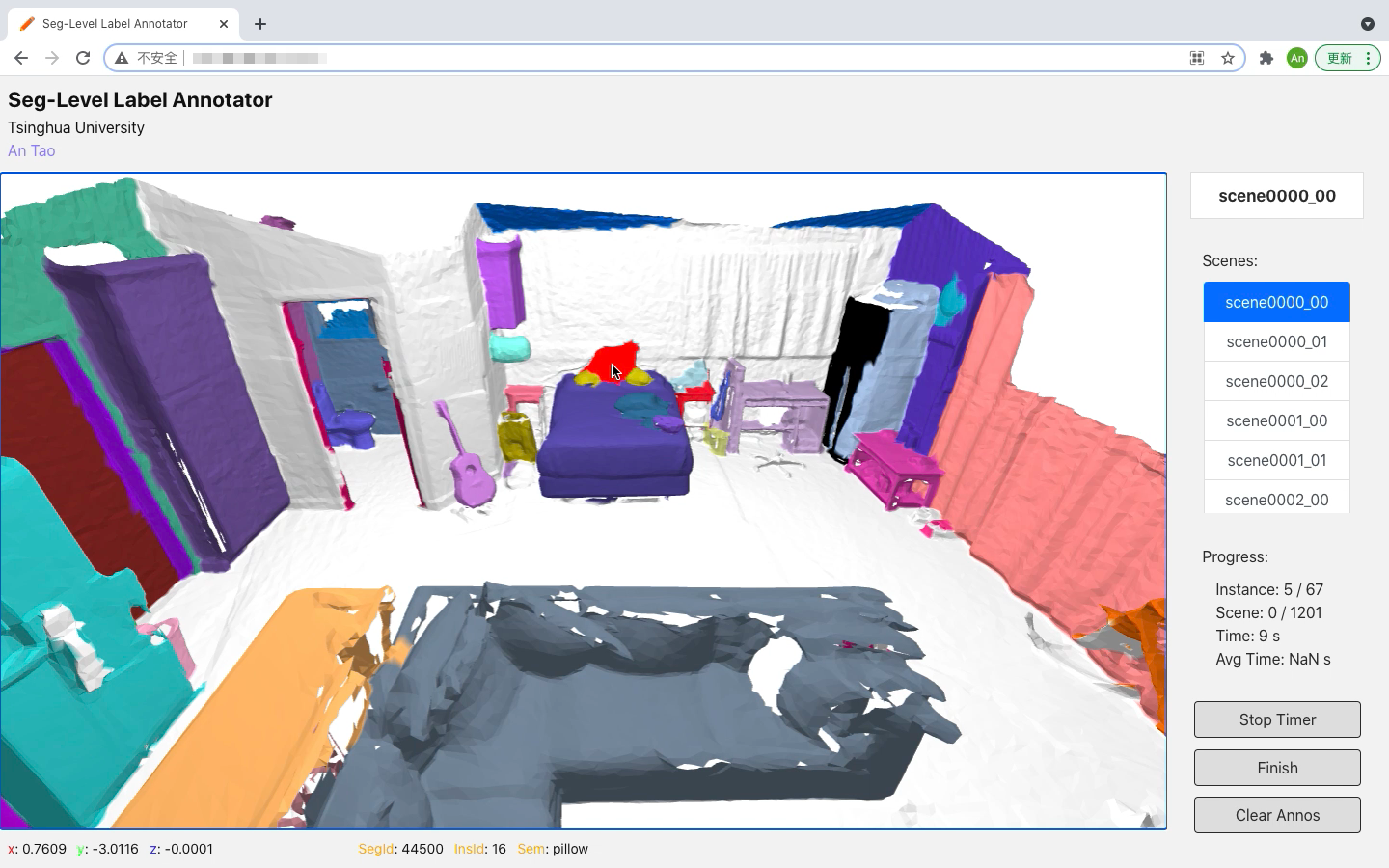}
		\end{minipage}
	}
	\subfigure[Annotation Results]{
		\label{anno_c}
		\begin{minipage}[b]{0.315\linewidth}
			\includegraphics[width=1\linewidth]{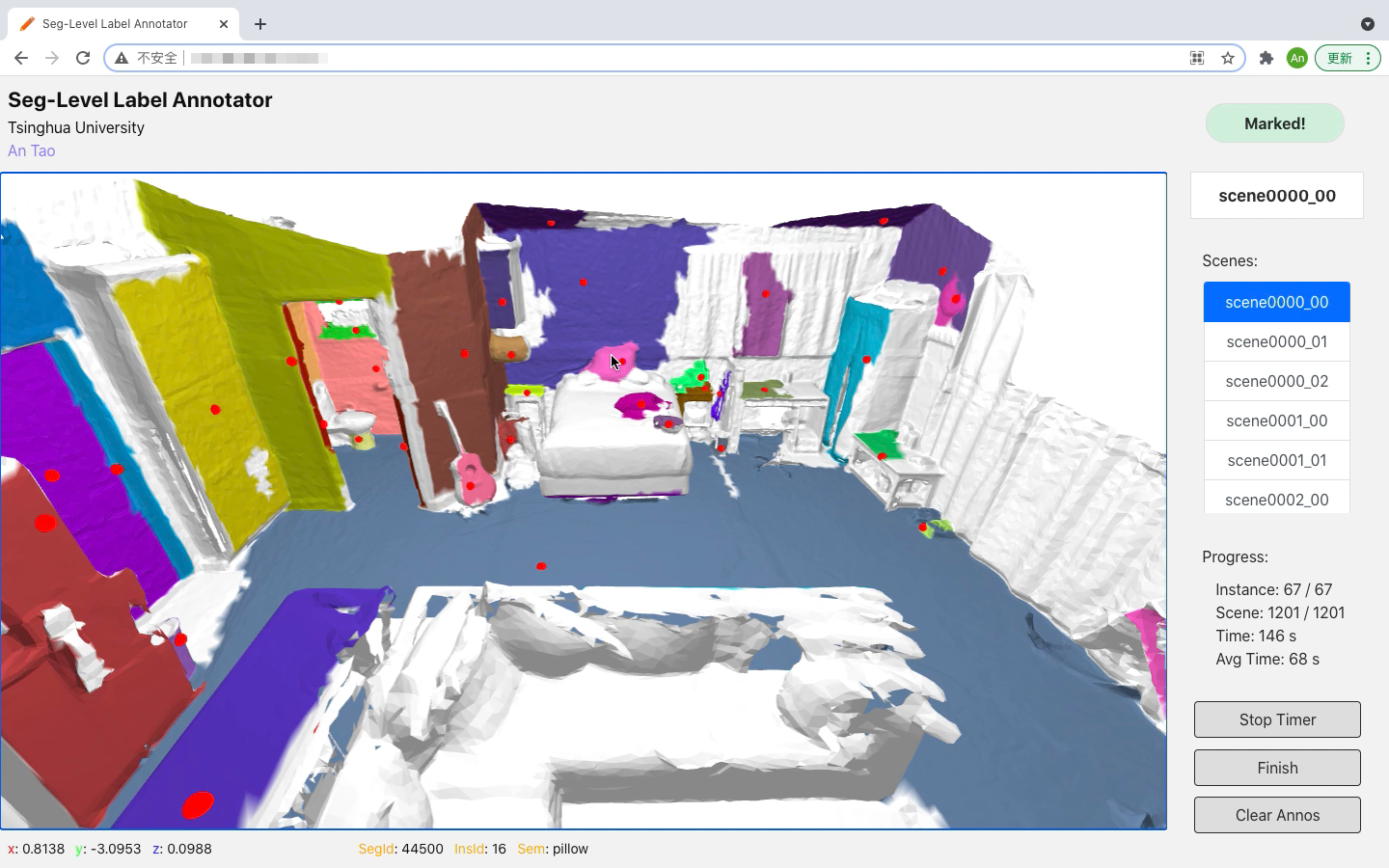}
		\end{minipage}
	}
		\vspace{5pt}
	\caption{The interfaces of our WebGL annotation tool. (a) We design an annotation interface that requires the annotator to label both the semantic class and the instance ID of each chosen segment. (b) Because scenes in ScanNet have ground-truth point-level labels, in this paper we choose to annotate our seg-level labels based on the ground-truth labels to reduce the annotation difficulty. (c) After annotation, the interface displays the annotation results including location annotations and seg-level labels.
	}
	\label{annotation}
%	\vspace{-5pt}
\end{figure*}

\subsection{Weakly-supervised Image Segmentation} Many works have been proposed for weakly-supervised image segmentation, where image-level supervision~\cite{pinheiro2015image,papandreou2015weakly,jing2019coarse,redondo2019learning,shi2016weakly} and bounding box supervision~\cite{dai2015boxsup,hsu2019weakly,khoreva2017simple} are two major lines in both instance and semantic segmentation. Image-level labels indicate the semantic classes appearing in an image, while bounding boxes further frame every instance with semantic classes. Weakly-supervised methods usually adopt a two-step process that first generates pseudo labels and then trains a supervised model treating these pseudo labels as ground truth. For image-level supervised segmentation, a common strategy is to train a classification model to recover class activation maps~\cite{zhou2016learning,wei2017object}. The predicted class activation maps are then used as 'seeds' for optimization methods that grow the coarse activation maps to larger pseudo segmentation maps. Some approaches additionally employ a class-agnostic saliency estimation model~\cite{chen2018encoder} to capture the objectness of pixels. In contrast, bounding boxes frame every object with semantic labels, alleviating the need to estimate class activation maps. Segmentation masks can further be refined by heuristic cues ~\cite{khoreva2017simple,rother2004grabcut} or mean-field inference~\cite{krahenbuhl2011efficient,papandreou2015weakly}. The refined masks are then adopted for segmentation model training. Some works additionally use EM\cite{dai2015boxsup,khoreva2017simple,papandreou2015weakly} for iterative refinement of the ground truth and model parameters.
There are also some other weak supervision forms~\cite{bearman2016s,lin2016scribblesup,li2019weaklier}. For example, Bearman~\textit{et al.}~\cite{bearman2016s} proposed point labels to annotate each object. Lin~\textit{et al.}~\cite{lin2016scribblesup} proposed scribble labels by dragging the cursor in the center of the objects.

\section{Proposed Approach}
In this section, we first describe the seg-level annotations for point cloud segmentation. Then, we detail our SegGroup network to generate point-level pseudo labels from the annotated seg-level labels on the training set. Finally, we introduce the network training and implementation details.

\subsection{Seg-level Annotation}
Following the labeling strategy of point-level labels in ScanNet~\cite{dai2017scannet}, we follow the over-segmentation results provided by the dataset, which is obtained by employing a normal-based graph cut method~\cite{felzenszwalb2004efficient,karpathy2013object} on the mesh. 
The scenes on the ScanNet dataset are provided as meshes. Each scene in the ScanNet dataset is reconstructed from an RGBD video stream that records every side of a room, and the mesh structure is universal in the reconstructed scenes of other indoor datasets. 
The vertices in the mesh are used as the input point cloud of our SegGroup network.
The results of over-segmentation remain unchanged in all the subsequent operations, including the manual annotation process and SegGroup network learning. 
%Although some adjacent similar parts that belong to different instances may be mistakenly segmented into one by the over-segmentation step, the annotators find that these effects are small and ignorable during the annotation process.

Unlike 2D images that may suffer from occlusions and lightning variances, 3D data structures usually have clear boundaries between different simple geometry parts thereby easier to perform over-segmentation. 
As segments are very small in this process, in most cases, each segment only contains one single object. We observe that for very few cases one segment may overlap different objects. Because the ground-truth strong labels of the ScanNet dataset are also annotated based on over-segmentation to accelerate the annotation process. Therefore, it is an intrinsic issue of this dataset that may have the minority of incorrect ground-truth labels, both in training and evaluation.

Before manual labeling started, we first generated various types of weak labels from ground-truth strong labels and conducted experiments to compare their performance in our manuscript in Table~\ref{label_type}. We find the performance mainly relies on the labeled segment sizes, i.e., the larger the better. According to our observation, the largest segment of each instance is usually the most central one, so we consider it can represent the location of the instance. During the annotation process, we ask the annotators to click on the point on the largest segment of the instance to show the location and consider the largest segment of an instance as the most representative segment. If it is hard to distinguish the largest segment (e.g. the sizes of some large segments for an instance are almost equal), we allow the annotators to click the point on the most central segment. With the help of over-segmentation, these location annotations are automatically extended into segments as seg-level labels. 

To annotate seg-level labels, we design a WebGL annotation tool in the browser. Fig.~\ref{annotation} shows the interface of our annotation tool, which includes a scene display window on the left and a control panel on the right. The annotator can rotate and pan the scene to browse and annotate seg-level labels by mouse clicking. Different from point-level labels that annotate every point in the scene, we click on one point per instance to indicate its location and give its semantic class. 

There are two modes to annotate the seg-level labels:

%\subsubsection{Annotating from Scratch}

\begin{figure*}[t]
	\begin{center}
		\includegraphics[width=1\linewidth, trim=20 110 570 50,clip]{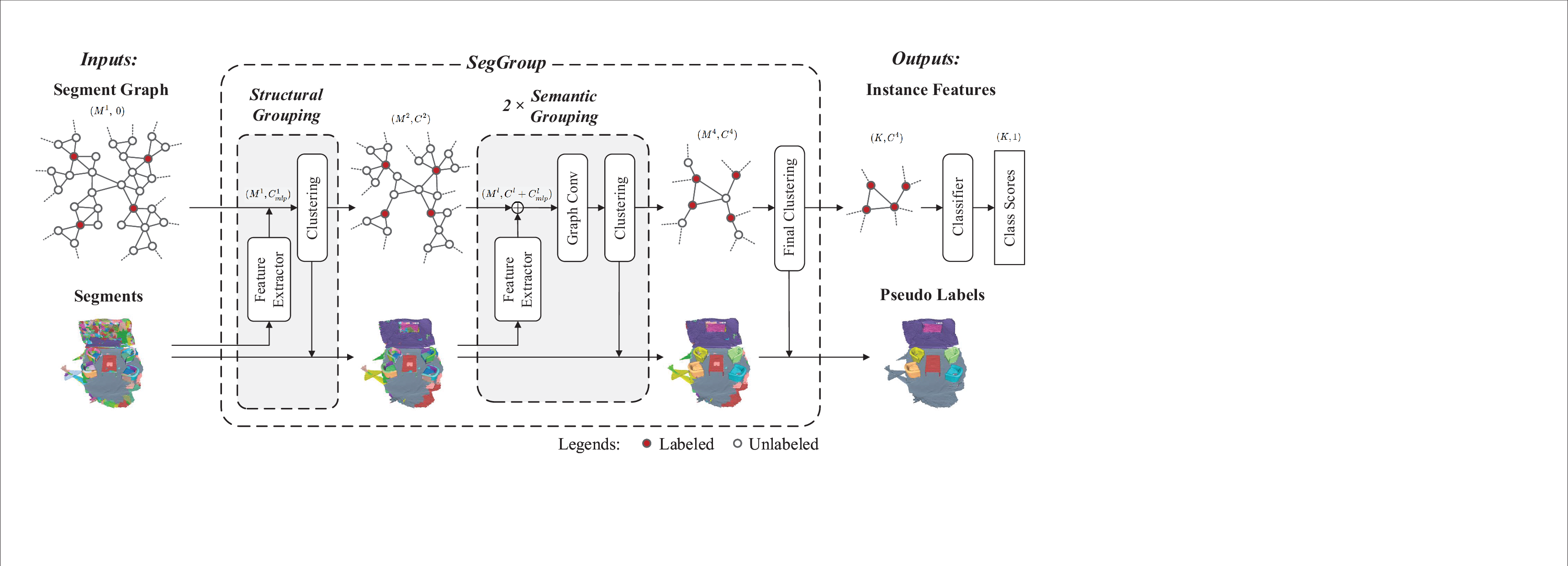}
	\end{center}
	\vspace{-5pt}
	\caption{The Structure of our SegGroup network. The inputs of the SegGroup network consist of a segment graph and the corresponding point cloud segments. Each node in the graph is attached with a feature vector, which describes the corresponding point cloud segment. Label information is extended from segments into nodes where the red color indicates labeled nodes. Edges in the graph denote the connected segments that are adjacent in the scene and the lengths of edges represent the similarity of segments. We design three grouping layers followed by a clustering step to group segments into instances by clustering nodes in the graph hierarchically. The classifier in the last part of the framework is used for network training.}
	%	\vspace{-5pt}
	\label{seggroup}
\end{figure*}

\textbf{Annotating from Scratch.} The annotation interface requires the annotator to label both the semantic class and the instance ID of the location of an instance (depicted in Fig.~\ref{anno_a}). In this annotation interface, the scene is displayed with original scanned colors at the beginning. The annotator needs to choose a semantic class before annotating the location of each instance. 
%If the next instance to annotate shares the same semantic class with the last annotated instance, the annotator only needs to click on \textquotedblleft Add\textquotedblright\, in the control panel to use the last chosen semantic class. To show the over-segmentation results to facilitate annotation, the segment corresponding to the mouse cursor is displayed in red. When an instance location is annotated, the color of the segment that corresponds to the instance location changes from red to a new color to indicate the segment is annotated. Different colors of the annotated segments indicate they belong to different instances. The mouse cursor location (XYZ) on the surface of the scene mesh and the ID of the segment that corresponds to the cursor location are shown at the bottom of the interface. 
In Fig.~\ref{anno_a}, the annotator is preparing to annotate the pillow. 

%\subsubsection{Annotating with GT Labels}

\textbf{Annotating with GT Labels.} Because scenes in ScanNet~\cite{dai2017scannet} dataset have ground-truth point-level labels, in this paper we choose to annotate our seg-level labels based on the ground-truth labels to reduce the annotation difficulty (depicted in Fig.~\ref{anno_b}). Compared with annotating from scratch in Fig.~\ref{anno_a}, in this annotation mode the annotator does not need to annotate the semantic class of each instance location. 
%In Fig.~\ref{anno_b}, different colors indicate different instances, and the white color indicates instances are labeled. The scene is displayed with non-white colors at the beginning of the annotation process. The annotator needs to annotate on every instance to make the scene become white in all areas. Same as the interface in Fig.~\ref{anno_a}, in Fig.~\ref{anno_b} the segment corresponding to the mouse cursor is displayed in red, and the status of the mouse cursor is shown at the bottom of the interface. Because the ground-truth labels of the scene are given in this annotation mode, the instance information of the segment that corresponds to the mouse cursor location is also displayed at the bottom of the interface. When an instance location is annotated by mouse clicking, the color of the segment that corresponds to the instance location changes from red to black. At the same time, the color of the instance last labeled turns white to indicate this instance is already labeled. In Fig.~\ref{anno_b}, the annotator has just annotated the cabinet and is preparing to annotate the pillow. The annotation status of Figs.~\ref{anno_a}-\ref{anno_b} is the same. Compared with annotating from scratch, in this annotation mode only when the annotator annotates a new instance, the annotation of the last instance is completed and the instance number is added in the annotation progress. Therefore, the instance number in Fig.~\ref{anno_b} is one less than the instance number in Fig.~\ref{anno_a}.

After annotation, the WebGL annotation tool provides an interface to display the annotation results (depicted in Fig.~\ref{anno_c}). The annotation results include location annotations and seg-level labels. In the display window, the positions of red balls indicate the location annotations of instances. For seg-level labels, different colors indicate they belong to different instances. The white areas of the scene are unlabeled. More details of the annotation tool can be accessed in \url{https://github.com/antao97/SegGroup.annotator}.

We calculate the average annotation time per scene based on our empirical timing results. In our work, because we only annotate the segment locations for each instance, the average instance annotation time (68s) is obtained by averaging the total annotation time for all 1,201 scenes. Our WebGL annotation tools can calculate the average annotation time per scene in real-time during the annotation process, as depicted in Figure~\ref{anno_c}. To simulate the standard annotation process from scratch in Fig.~\ref{anno_a}, we need to add the semantic annotation time. The average time to annotate a semantic class (1.5s) is obtained by testing the annotation time empirically on 10 scenes. Because the average instance number of each scene is 32, we can finally derive our total annotation time for a scene by $68 + 32\times 1.5 = 116$s (1.93 minutes).

\begin{figure*}[t]
	%		\vspace{-10pt}
	\begin{center}
		\includegraphics[width=0.95\linewidth, trim=50 110 600 130,clip]{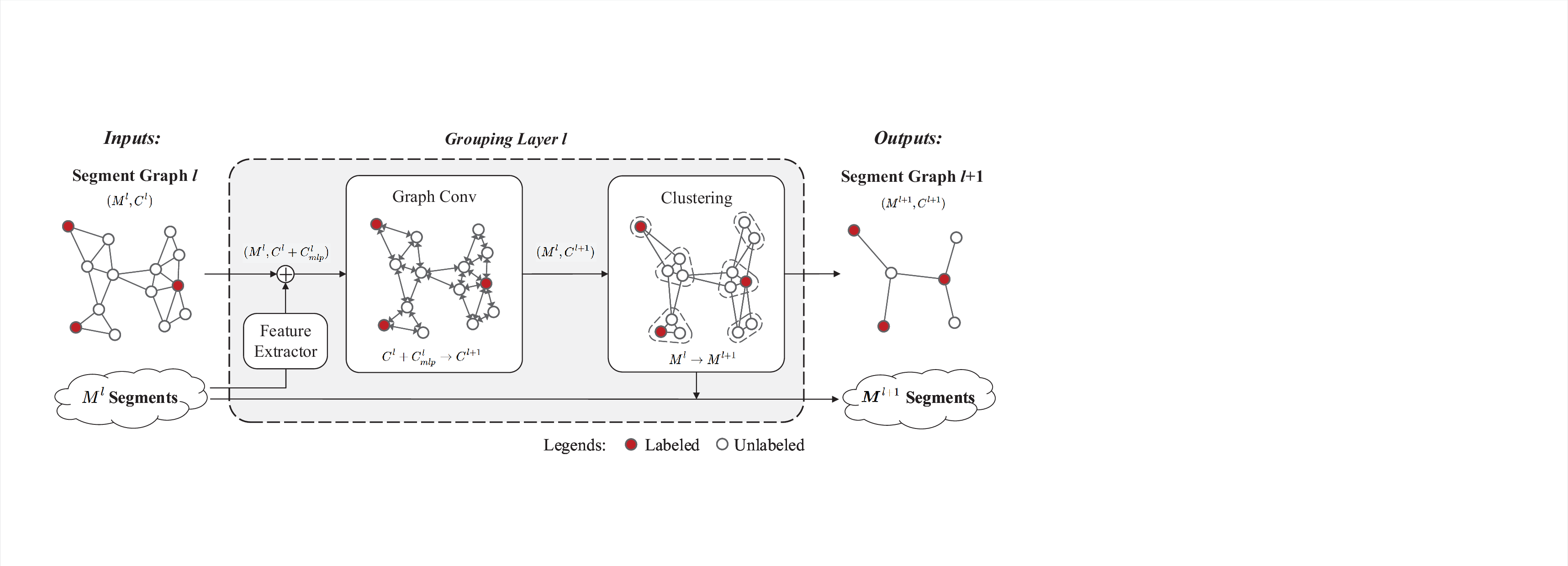}
	\end{center}
	\vspace{-5pt}
	\caption{The Structure of the semantic grouping layer. 
		The inputs consist of a segment graph in layer $l$ as well as the corresponding point cloud segments. Same as the settings in Fig.~\ref{seggroup}, each node in the graph represents a unique point cloud segment. 
		The node features are first concatenated with segment features from point cloud segments by a feature extractor, and then updated by a graph convolution network. Finally, a node clustering algorithm groups similar nodes into new nodes. With the clustering result, similar segments are merged into large segments. }
	\label{grouping_layer}
	%	\vspace{-5pt}
\end{figure*}

\subsection{SegGroup}
\label{seggroup_sec}
To learn point cloud segmentation models with seg-level labels on the training set, we propose a two-stage approach. We first design a segment grouping network (SegGroup) to generate pseudo labels for the remaining unlabeled points on the training set. Then, we adopt an existing point-level supervised point cloud segmentation model (such as PointGroup~\cite{jiang2020pointgroup}) to consume the generated point-level pseudo labels for standard training and evaluate its performance on the testing set. The two stages are trained separately.

The structure of our SegGroup network is shown in Fig.~\ref{seggroup}. To fully utilize the locations of instances in 3D scene to generate pseudo labels, we assume that all segments of a single instance are interconnected so that segments can be gathered into instances according to their neighbor relationship under the guidance of instance locations. With the mesh structure of the scene data, we consider two segments to be neighbors if there exists an edge connecting two vertices belonging to two segments. Therefore, given an over-segmented point cloud scene, we build a segment graph where each node denotes a unique segment and the edge shows the adjacency between two neighboring segments in the scene. We extend the label information of each segment to the corresponding node in the graph, where the red color indicates labeled nodes. As the labeled nodes show the locations of instances, unlabeled nodes of the same instance are gradually grouped into labeled nodes through edges in the graph with our SegGroup network. The final outputs of the SegGroup are one node for each instance and all nodes are labeled with semantic classes and different instance IDs. When the nodes are merged, the segments are also merged into larger segments until all unlabeled segments are merged into nearby labeled segments. All points in the 3D scene are therefore labeled with semantic classes and instance IDs, and we take these labels as point-level pseudo labels. The additional classifier in the last part of the framework is used for network training.
%Each node is also attached with a feature vector as node feature, and at the beginning there are no features. 

We design three grouping layers followed by a final clustering process to group segments into instances by clustering nodes in the graph hierarchically. The first layer is a structural grouping layer, whose objective is to group similarly structured segments into one segment to reduce the computational costs of the subsequent grouping layers. The second and the third layer are semantic grouping layers. Because the semantic and structural grouping layers are almost the same except that the semantic grouping layer has an additional graph convolution network, we only introduce the semantic grouping layer in the following content. 
%Different from structural grouping layer, a graph convolution network is used in semantic grouping layer to update node features with a weighted sum of its neighbors. Note that in all layers we first concatenate node features with segment features extracted from point cloud to maintain node feature discriminative. The final grouping stage group all unlabeled nodes into nearby relevant nodes to generate instance proposals. 
%Finally, we obtain a graph containing $K$ node features and the corresponding $K$ point cloud instances. For each feature, we use a classifier to obtain a score of labeled semantic class for back-propagation. After the training process of SegGroup network, the output point cloud instances serve as point-level pseudo labels for strong supervised instance and semantic point cloud segmentation. 

The structure of the semantic grouping layer is shown in Fig.~\ref{grouping_layer}. The inputs of the $l$-th layer consist of a segment graph with $M^l$ nodes as well as its corresponding $M^l$ point cloud segments. Each node is represented by a $C^l$ dimensional vector which is considered as the node feature. The output of the grouping layer is a new graph with $M^{l+1}$ nodes as well as its corresponding $M^{l+1}$ point cloud segments, where $M^{l+1} \leq M^l$. The node features of the output new graph are in $C^{l+1}$ dimensional.
%Same as the settings in Figure~\ref{seggroup}, the red color indicates labeled nodes in Figure~\ref{grouping_layer}. 

In the following content, we introduce the specific blocks of the semantic grouping layer.

\textbf{Feature Extractor.} We adopt a shared EdgeConv~\cite{wang2019dynamic} network to obtain $C^l_{mlp}$ dimensional segment features for each input point cloud segment of the grouping layer individually, which serves as a local feature learning module to extract semantic information. 
%The EdgeConv network is depicted as \textquotedblleft Feature Extracor\textquotedblright~in Fig.~\ref{seggroup}-\ref{grouping_layer}. 
During the forward-propagation process of the SegGroup network, segments are gradually merged, so that the receptive field of the feature extractor in the deeper grouping layer also becomes larger to extract more macroscopic information. The input $C^l$ dimensional node features are concatenated with their corresponding newly extracted $C^l_{mlp}$ dimensional segment features to form new node features in $C^l+C^l_{mlp}$ dimension.
%At each grouping layer, we first concatenate the node features with the segment features extracted from point cloud segments. 

\textbf{Graph Convolution.} We update the node features with a graph convolution network (GCN)~\cite{kipf2016semi}.
%, which is depicted as \textquotedblleft Graph Conv\textquotedblright~in Fig.~\ref{seggroup}-\ref{grouping_layer}. 
%By clustering the similar segments into new segments, the grouping layer outputs a new segment graph at layer $l+1$ as well as the corresponding $M^{l+1}$ point cloud segments. 
%The inputs are a segment graph in layer $l$ and $M^l$ point cloud segments. The segment graph in layer $l$ has $M^l$ nodes, each of which represents a unique point cloud segment and has a $C^l$-dim feature vector. Edges in the graph denote the connected segments are adjacent in the point cloud, and edge lengths represent the similarity of segments. In the grouping stage, the feature of each segment is first concatenated with a new $C^l_{mlp}$-dim feature vector extracted from the corresponding point cloud segment by a MLP. Then, a graph convolution network calculates the feature similarities between connected nodes as propagation weights and updates segment features into $C^{l+1}$-dim. Finally, the clustering algorithm groups segments into new segments according to their features. The outputs of the grouping stage are a new segment graph of layer $l+1$ and $M^{l+1}$ point cloud segments.
Through the neighbor relationship of nodes in the graph, the goal of GCN is to semantically narrow down the difference between nodes belonging to the same instance and extend the difference between nodes belonging to different instances. 
%Through the neighbor relationship of nodes in the graph, similar node features of adjacent nodes can be gathered more closer in the feature space by GCN. The differences between nodes belonging to different instances are therefore expended. 
Given a node feature $\vec{h}_i^l$ and its adjacent neighbor node feature $\vec{h}_j^l$ in the $l$-th grouping layer, the similarity coefficient $e_{ij}^l$ is computed according to the distance between $\vec{h}_i^l$ and $\vec{h}_j^l$ as
\begin{equation}
e_{ij}^l = {\rm exp}(-\lambda\lVert\vec{h}_i^l-\vec{h}_j^l\rVert_2),
\end{equation}
where $\lambda$ is a positive parameter to control the slope. A small distance indicates similar node features, which leads to a large coefficient. If $i=j$, the coefficient $e_{ii}^l=1$. To make coefficients easily comparable across different nodes, we normalize them across all choices of $j$ as 
\begin{equation}
a_{ij}^l = \frac{e_{ij}^l}{e_{ii}^l+\sum_{k\in\mathcal{N}_i^l}e_{ik}^l},
\end{equation}
where $\mathcal{N}_i^l$ is the neighborhood of node $i$ in the graph. 
The normalized similarity coefficients are used to compute a linear combination of adjacent segment features. After a shared linear transformation parametrized by a weight matrix $\mathbf{W}^l$ and a nonlinearity $\sigma$, the output node feature $\vec{h}_i^{l'}$ is finally computed as
\begin{equation}
\vec{h}_i^{l'} = \sigma\left(a_{ii}^l\mathbf{W}^l\vec{h}_i^l+\sum_{k\in\mathcal{N}_i^l}a_{ik}^l\mathbf{W}^l\vec{h}_k^l\right).
\end{equation}
In grouping layer $l$, the dimension of $\vec{h}_i^l$ is $C^{l}+C^l_{mlp}$, and the dimension of $\vec{h}_i^{l'}$ is $C^{l+1}$.

\begin{algorithm}[t]
	\caption{Clustering in Layer $l$}
	\hspace*{0.02in} {\bf Input:}
	Nodes $\{n_i^l\}_{i=1}^{M^l}$, node labels $\{y_i^l\}_{i=1}^{M^l}$, \\
	\hspace*{0.2in} node features $\{\vec{h}_i^{l'}\}_{i=1}^{M^l}$,	edges $\{o_s^l\}_{s=1}^{P^l}$, \\
	\hspace*{0.2in} distance threshold $e_{\tau}^l$.\\
	\hspace*{0.02in} {\bf Output:} 
	%	Clusters $\{C_i\}_{i=1}^{M^{l+1}}$, cluster labels $\{z_i\}_{i=1}^{M^{l+1}}$, cluster features $\{z_i\}_{i=1}^{M^{l+1}}$.
	Nodes $\{n_i^{l+1}\}_{i=1}^{M^{l+1}}$, node labels $\{y_i^{l+1}\}_{i=1}^{M^{l+1}}$, \\
	\hspace*{0.2in} node features $\{\vec{h}_i^{l+1}\}_{i=1}^{M^{l+1}}$,	edges $\{o_s^{l+1}\}_{s=1}^{P^{l+1}}$.
	\begin{algorithmic}[1]
		\For{every node $n_i^l$}
		\State $C_i^l \leftarrow \{n_i^l\}$ ~~~~~~~~~// initialize clusters
		\State $z_i^l \leftarrow y_i^l$ ~~~~~~~~~~~~// initialize cluster labels
		%		\State $\vec{v}_i^{\,l} = \vec{h}_i^{l'}$ ~~~~~~~~// initialize cluster features
		\EndFor
		\For{every edge $o_s^l=(n_a^l, n_b^l)$} 
		\State Find $n_a^l\in C_p^l$, $n_b^l\in C_q^l$
		\If{$C_p^l == C_q^l$} 
		\State Continue ~~~~~~~// belong to same cluster
		\EndIf
		\If{$z_p^l\neq {\rm None}$ and $z_q^l\neq {\rm None}$} 
		\State Continue ~~~~~~~~~~~~~~// both clusters are labeled 
		\EndIf
		\If{${\rm dist}(\vec{h}_a^{l'}, \vec{h}_b^{l'}) < e_{\tau}^l$}
		\State $C_p^l \leftarrow C_p^l\cup C_q^l$ ~~~~~// merge clusters
		\State Delete $C_q^l$ 
		\If{$z_q^l\neq {\rm None}$}
		\State $z_p^l \leftarrow z_q^l$ ~~~~~~~~~~~~~~// update cluster label
		\EndIf
		\State Delete $z_q^l$ 
		\EndIf
		\EndFor
		\State Obtain $M^{l+1}$ clusters and $M^{l+1}$ labels
		\State Sort clusters and labels into $\{C_i^l\}_{i=1}^{M^{l+1}}$ and $\{z_i^l\}_{i=1}^{M^{l+1}}$
		\For{every cluster $C_i^l$}
		\State $n_i^{l+1} \leftarrow C_j^l$~~~~~~~~~~~~~~~~~~~~~~~~~~~// get node
		\State $y_i^{l+1} \leftarrow z_i^l$ ~~~~~~~~~~~~~~~~~~~~~~~~~~~// get node label
		\State $\vec{h}_i^{l+1} \leftarrow {\rm maxpool}(\{\vec{h}_k^{l'}\}_{n_k^l\in C_i^l})$ ~~~// get node feature
		\EndFor
		%		\State Obtain new nodes $\{j\}_{j=1}^{M^{l+1}}$
		%		\State Obtain new node labels $\{y_j^{l+1}\}_{j=1}^{M^{l+1}}$
		%		\State Obtain new node features $\{\vec{h}_j^{l+1}\}_{j=1}^{M^{l+1}}$
		
		\For{every edge $o_s^l=(n_a^l, n_b^l)$} 
		\State Find $n_a^l\in C_p^l$, $n_b^l\in C_q^l$
		\If{$C_p^l \neq C_q^l$} 
		\State $o_s^{l+1} \leftarrow (n_p^{l+1}, n_q^{l+1})$ ~~~~~~~~~~~~~~// get edge
		\EndIf
		\EndFor
		\State Obtain $P^{\,l+1}$ edges
		\State Sort edges into $\{o_s^{l+1}\}_{s=1}^{P^{\,l+1}}$
		\State \Return $\{n_i^{l+1}\}_{i=1}^{M^{l+1}}$, $\{y_i^{l+1}\}_{i=1}^{M^{l+1}}$, $\{\vec{h}_i^{l+1}\}_{i=1}^{M^{l+1}}$,
	\end{algorithmic}
	\hspace*{0.2in} $\{o_s^{l+1}\}_{s=1}^{P^{\,l+1}}$ ~~~~~~~~~~~~~~// return results for next layer
	\label{algorithm}
\end{algorithm}

\begin{algorithm}[t]
	\caption{Final Clustering}
	\hspace*{0.02in} {\bf Input:}
	Nodes $\{n_i^l\}_{i=1}^{M^l}$, node labels $\{y_i^l\}_{i=1}^{M^l}$, \\
	\hspace*{0.2in} node features $\{\vec{h}_i^{l}\}_{i=1}^{M^l}$,	edges $\{o_s^l\}_{s=1}^{P^l}$. \\
	\hspace*{0.02in} {\bf Output:} 
	%	Clusters $\{C_i\}_{i=1}^{M^{l+1}}$, cluster labels $\{z_i\}_{i=1}^{M^{l+1}}$, cluster features $\{z_i\}_{i=1}^{M^{l+1}}$.
	Nodes $\{n_i^l\}_{i=1}^{K}$, node labels $\{y_i^{l}\}_{i=1}^{K}$, \\
	\hspace*{0.2in} node features $\{\vec{h}_i^{l}\}_{i=1}^{K}$, edges $\{o_s^l\}_{s=1}^{Q}$.
	\begin{algorithmic}[1]
		\While{exist $y_i^l == {\rm None}$}
		\For{every node $n_i^l$ that $y_i^l == {\rm None}$}
		\State Find the neighborhood $\mathcal{N}_i^l$ of node $n_i^l$ by edges
		\State Find node $n_k^l \in \mathcal{N}_i^l$ that minimize ${\rm dist}(\vec{h}_i^{l}, \vec{h}_k^{l})$
		\State $n_k^l \leftarrow \{n_i^l, n_k^l\}$
		\State Delete node $n_i^l$
		\State Delete node label $y_i^l$
		\State $\vec{h}_k^{l} \leftarrow {\rm maxpool}(\{\vec{h}_i^{l}, \vec{h}_k^{l'}\})$
		\State Delete node feature $\vec{h}_i^{l}$
		\State Delete edge $o_s^l = (i,k)$
		\EndFor
		\EndWhile
		\State Obtain $K$ nodes, node labels, and node features
		\State Sort into $\{n_i^l\}_{i=1}^{K}$, $\{y_i^{l}\}_{i=1}^{K}$, $\{\vec{h}_i^{l}\}_{i=1}^{K}$
		\State Obtain $Q$ edges
		\State Sort into $\{o_s^{l}\}_{s=1}^{P^{l'}}$
		\State \Return $\{n_i^l\}_{i=1}^{K}$, $\{y_i^{l}\}_{i=1}^{K}$, $\{\vec{h}_i^{l}\}_{i=1}^{K}$, $\{o_s^{l}\}_{s=1}^{Q}$ 
	\end{algorithmic}
	\label{algorithm2}
\end{algorithm}

\textbf{Clustering.} We further design a straightforward but effective node clustering algorithm to group similar neighboring nodes into one node. Given a grouping layer $l$, for all neighboring pairwise nodes, if they do not belong to different instances and the distance between node features is below a given threshold $e_{\tau}^l$, we merge the two nodes into a new node. Finally, the algorithm produces $M^{l+1}$ nodes from $M^{l}$ nodes in grouping layer $l$. The node features of the new nodes are obtained by a max-pooling operation on their merged node features.

Algorithm~\ref{algorithm} shows the detailed procedure to produce $M^{l+1}$ nodes from $M^{l}$ nodes in grouping layer $l$. Given a graph that has $M^l$ nodes, we first initialize each node with a separate cluster and also assign the label information (labeled or unlabeled) to the clusters. Then, we gradually merge clusters according to conditions. More specifically, for each edge in the graph, we first check whether the connected two nodes are in the same cluster or their corresponding clusters are both labeled. If neither of the two conditions is true, the two clusters do not belong to different instances and then we check whether the distance between the two node features is smaller than a threshold $e_{\tau}^l$. If the distance condition is met, we merge the two clusters and update label information. After finishing the edge traversal, we obtain $M^{l+1}$ clusters and $M^{l+1}$ corresponding labels. Then, we convert these clusters into new nodes and extend the label information to them. The node feature of each new node is obtained by a max-pooling operation on old node features of all nodes in each cluster. Before the max-pooling operation, the node features and node-wise distances are unchanged during the clustering process. Finally, we remove the edges inside each cluster to obtain a new graph. The node features are uniformly updated at the end of the process.

The label propagation process of our method considers the instance IDs of labeled segments by restraining the merging of two clusters with different instance IDs (lines 8-9 in Algorithm~\ref{algorithm}). There may exist issues in the boundaries between two neighbor instances of the same semantic class, but we find this circumstance is very rare in 3D scenes.
If this circumstance happens, the boundaries between the two instances are usually very clear. The segments of different instances near the boundaries cannot be easily merged due to geometry differences. 
%Our method fully considers the geometry information during label propagation in the Structural Grouping stage, which is the first grouping stage in our framework.

After the three grouping layers in Fig.~\ref{seggroup}, most of the nodes are merged. For the few remaining unlabeled nodes, we further perform the below node clustering algorithm to group all unlabeled nodes into nearby relevant nodes to generate final instance proposals. 

\textbf{Final Clustering.} Different from the node clustering algorithm in the three grouping layers, this node clustering algorithm traverses all unlabeled nodes. This algorithm adopts a greedy strategy that obtains a locally optimal solution by merging each unlabeled node into the most similar neighbor node in the graph by comparing distances between node features. Finally, all the nodes in the graph are labeled and each instance in the 3D scene corresponds to a unique node. The point cloud segments become our propagated instance proposals which are considered as point-level pseudo labels. 

Algorithm~\ref{algorithm2} shows the detailed procedure to produce $K$ nodes from $M^{l}$ nodes after all grouping layers, where $K$ is the number of instances in the 3D scene. For each unlabeled node, this algorithm merges it into the most similar neighbor node in the graph by comparing distances between node features. At each step when two nodes are merged, the label information is updated and the new node feature is computed by a max-pooling operation on the two old node features. The corresponding edge is also removed to form a new graph. In the point cloud scene, the two corresponding segments are also merged into one. The clustering process continues until no unlabeled nodes exist. 

Although Algorithm~\ref{algorithm2} depends on the order of the nodes, according to our design most of the nodes are meaningfully clustered by Algorithm~\ref{algorithm} before the final node clustering process. The goal of the final node clustering in Algorithm~\ref{algorithm2} acts as a cleaning-up role to quickly combine the few remaining unlabeled nodes into nearby related nodes. The order of the nodes only has a minor effect in Algorithm~\ref{algorithm2}. 
To validate the stability of our method, we randomly shuffle the order of nodes 3 times and find the floating range of the mIoU of generated pseudo labels is within 0.5\% in semantic mIoU.

After the final clustering process, we can obtain point-level pseudo labels of an input scene. The nodes of the final graph in Fig.~\ref{grouping_layer} have different instance IDs. In the next section, we show how to train the SegGroup network in order to get better pseudo labels.

%we group all unlabeled nodes in the graph into nearby relevant nodes to generate instance proposals. Similar to the clustering stage in the grouping layer, we first initialize the nodes with separate clusters and assign labels for them. Then, we search for the most similar neighboring cluster for each unlabeled cluster and merge them into a larger one. After traversing until all the unlabeled nodes are grouped into labeled clusters, we obtain a graph with new nodes and features. Finally, all point cloud segments are grouped into labeled instances according to the clustering results, so that we assign all the points with pseudo labels.

\subsection{Network Training}
The SegGroup network outputs a graph containing $K$ nodes, each of which is attached with a label and a node feature. Because these $K$ nodes correspond to $K$ different instances, we extend the label information into instances and consider the node features as instance features. To train the SegGroup network, we adopt a classifier network to obtain a score of labeled semantic class for each instance. 
A cross-entropy loss is further computed for backward-propagation. In essence, our network training scheme alternates between two steps.
\begin{enumerate}
	\item Forward-propagation. With the network parameters fixed, the SegGroup propagates labels to unlabeled segments as pseudo labels by clustering. The output instance features are further processed by the classifier to obtain semantic scores. 
	
	\item Backward-propagation. With the pseudo labels fixed, we use the cross-entropy loss computed by the semantic scores to optimize the parameters of the SegGroup and the classifier. As the network parameters are optimized, the instance features which are gathered by the node features in all grouping layers can capture more semantic information. 
\end{enumerate}

Similar ideas of this EM-like algorithm are widely-used to effectively refine the generated pseudo labels in weakly- and semi-supervised learning, such as \cite{dai2015boxsup,lin2016scribblesup,papandreou2015weakly}. 

Although the clustering algorithm may make mistakes and spread to the ambiguous parts with the hierarchical grouping operations, the generated pseudo labels are relatively accurate around the locations of instances that are indicated by seg-level labels. During the training process, the feature extractor and GCN can gradually learn semantically related features for each instance according to the 3D structures around the instance locations. As the features of nodes belonging to the same instance contain more semantic information and become closer in feature space, the clustering algorithm can gather them more correctly with the help of the given instance location and the SegGroup can output better pseudo labels. After such a virtuous cycle in network training, we finally obtain the well-trained pseudo labels and they are used in the strong supervised instance and semantic point cloud segmentation. 

\begin{table}[t]
	\footnotesize
	\caption{The dimensions of the features in our architecture.}
	\begin{center}
		\begin{tabular}{l|c||l|c}
			\toprule
			Feature & Dimension & Feature & Dimension\\
			\midrule
			~~$C^1$ & 0 & ~~$C^1_{mlp}$ & 128 \\
			~~$C^2$ & 128 & ~~$C^2_{mlp}$ & 64 \\
			~~$C^3$ & 192 & ~~$C^3_{mlp}$ & 64 \\
			~~$C^4$ & 256 && \\
			\bottomrule
		\end{tabular}
	\end{center}
	\vspace{-5pt}
	\label{para}
\end{table}

\begin{table*}[t]
	\footnotesize
	\caption{The class-specific semantic IoUs (\%) of the generated pseudo labels on ScanNet training set. }
	\vspace{-7pt}
	{\scriptsize
		\begin{center}
			\begin{tabular}{l|c|c|m{0.2cm}<{\centering} m{0.2cm}<{\centering} m{0.2cm}<{\centering} m{0.2cm}<{\centering} m{0.2cm}<{\centering} m{0.2cm}<{\centering} m{0.2cm}<{\centering} m{0.2cm}<{\centering} m{0.2cm}<{\centering} m{0.2cm}<{\centering} m{0.2cm}<{\centering} m{0.2cm}<{\centering} m{0.2cm}<{\centering} m{0.2cm}<{\centering} m{0.2cm}<{\centering} m{0.2cm}<{\centering} m{0.2cm}<{\centering} m{0.2cm}<{\centering} m{0.2cm}<{\centering} m{0.4cm}<{\centering}|m{0.4cm}<{\centering}}
				\toprule
				Method & Label& Anno. Time & ~~\rotatebox{90}{Wall} & ~~\rotatebox{90}{Floor} & ~~\rotatebox{90}{Cab.} & ~~\rotatebox{90}{Bed} & ~~\rotatebox{90}{Chair} & ~~\rotatebox{90}{Sofa} & ~~\rotatebox{90}{Table} & ~~\rotatebox{90}{Door} & ~~\rotatebox{90}{Wind.} & ~~\rotatebox{90}{Bshf.} & ~~\rotatebox{90}{Pic.} & ~~\rotatebox{90}{Cntr.} & ~~\rotatebox{90}{Desk} & ~~\rotatebox{90}{Curt.} & ~~\rotatebox{90}{Fridg.} & ~~\rotatebox{90}{Shwr.} & ~~\rotatebox{90}{Toil.} & ~~\rotatebox{90}{Sink} & ~~\rotatebox{90}{Bath.} & \rotatebox{90}{Ofurn.} & \rotatebox{90}{Mean}\\
				\midrule
				MPRM~\cite{wei2020multi} & Scene & 0.25 min & 47.3 & 41.1 & 10.4 & 43.2 & 25.2 & 43.1 & 21.9 & ~9.8 & 12.3 & 45.0 & ~9.0 & 13.9 & 21.1 & 40.9 & ~1.8 & 29.4 & 14.3 & ~9.2 & 39.9 & 10.0 & 24.4 \\
				MPRM~\cite{wei2020multi} & Subcloud & 3 min& 58.0 & 57.3 & 33.2 & \textbf{71.8} & 50.4 & \textbf{69.8} & 47.9 & 42.1 & 44.9 & \textbf{73.8} & 28.0 & 21.5 & 49.5 & \textbf{72.0} & 38.8 & 44.1 & 42.4 & 20.0 & 48.7 & 34.4 & 47.4\\
				\midrule
				Layer 1 & Seg & 1.93 min & 37.4 & 51.5 & 30.5 & 20.3 & 18.7 & 11.3 & 35.0 & 38.5 & 11.6 & ~9.2 & \textbf{47.6} & 36.4 & 22.0 & 5.5 & 25.3 & 14.0 & 13.5 & 21.2 & 21.0 & 20.2 & 24.5\\
				Layer 2 & Seg & 1.93 min & 65.5 & 81.4 & 44.1 & 35.0 & 31.5 & 22.3 & 49.0 & 61.4 & 36.3 & 17.6 & 43.6 & 49.1 & 36.7 & 56.4 & 41.4 & 69.1 & 22.0 & 27.8 & 25.2 & 37.0 & 42.6\\
				Layer 3 & Seg & 1.93 min & 67.8 & 80.1 & 50.1 & 37.9 & 43.7 & 32.0 & 51.8 & 64.2 & 43.2 & 27.4 & 43.5 & 49.2 & 39.7 & 61.2 & 49.3 & 70.0 & 31.0 & 36.1 & 29.6 & 48.6 & 47.8\\
				Layer 4 & Seg & 1.93 min & 67.9 & 80.1 & 50.9 & 38.1 & 51.2 & 33.5 & 54.5 & 64.3 & 43.7 & 28.4 & 44.9 & 49.3 & 40.5 & 61.4 & 51.5 & 70.0 & 36.0 & 42.1 & 30.3 & 51.2 & 49.5\\
				\midrule
				SegGroup & Seg & 1.93 min & \textbf{71.0} & \textbf{82.5} & \textbf{63.0} & 52.3 & \textbf{72.7} & 61.2 & \textbf{65.1} & \textbf{66.7} & \textbf{55.9} & 46.3 & 42.7 & \textbf{50.9} & \textbf{50.6} & 67.9 & \textbf{67.3} & \textbf{70.3} & \textbf{70.7} & \textbf{53.1} & \textbf{54.5} & \textbf{63.7} & \textbf{61.4}\\
				\bottomrule
			\end{tabular}
		\end{center}
	}
	\label{sem_iou}
\end{table*}

\begin{figure*}[t]
	\centering
	\vspace{10pt}
	\subfigure[Point Cloud Scenes]{
		\begin{minipage}[b]{0.23\linewidth}
			\includegraphics[width=1\linewidth, trim=0 0 0 0,clip]{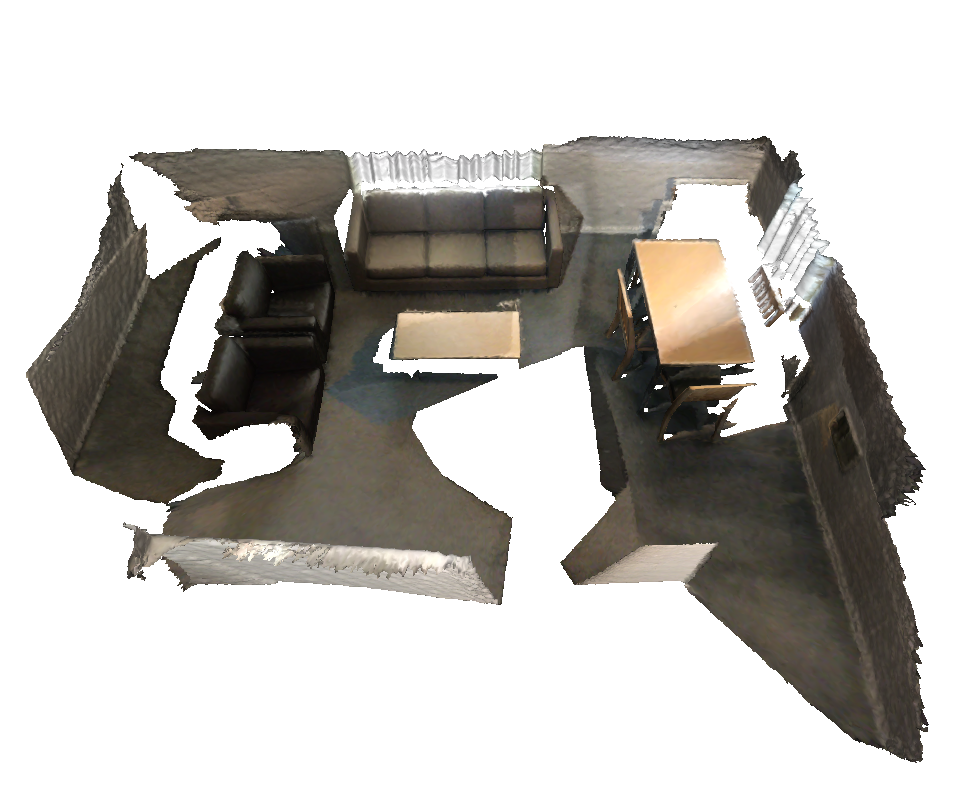}
			\vspace{7pt}
			\includegraphics[width=1\linewidth, trim=0 0 0 0,clip]{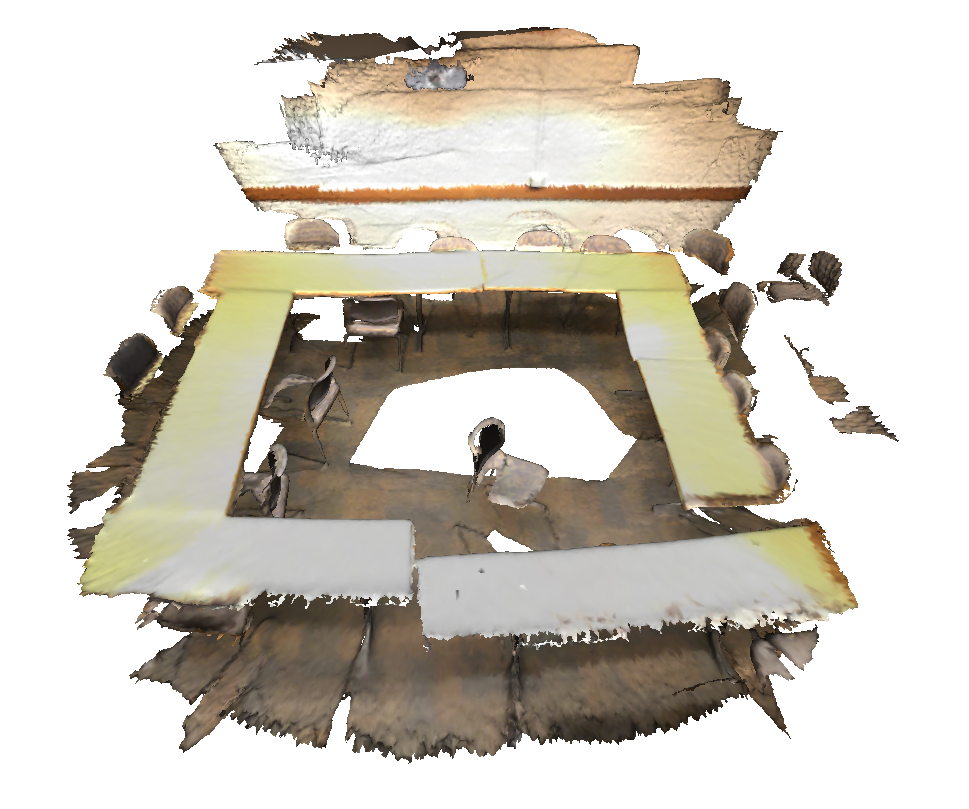}
			\vspace{7pt}
			\includegraphics[width=1\linewidth, trim=0 0 0 0,clip]{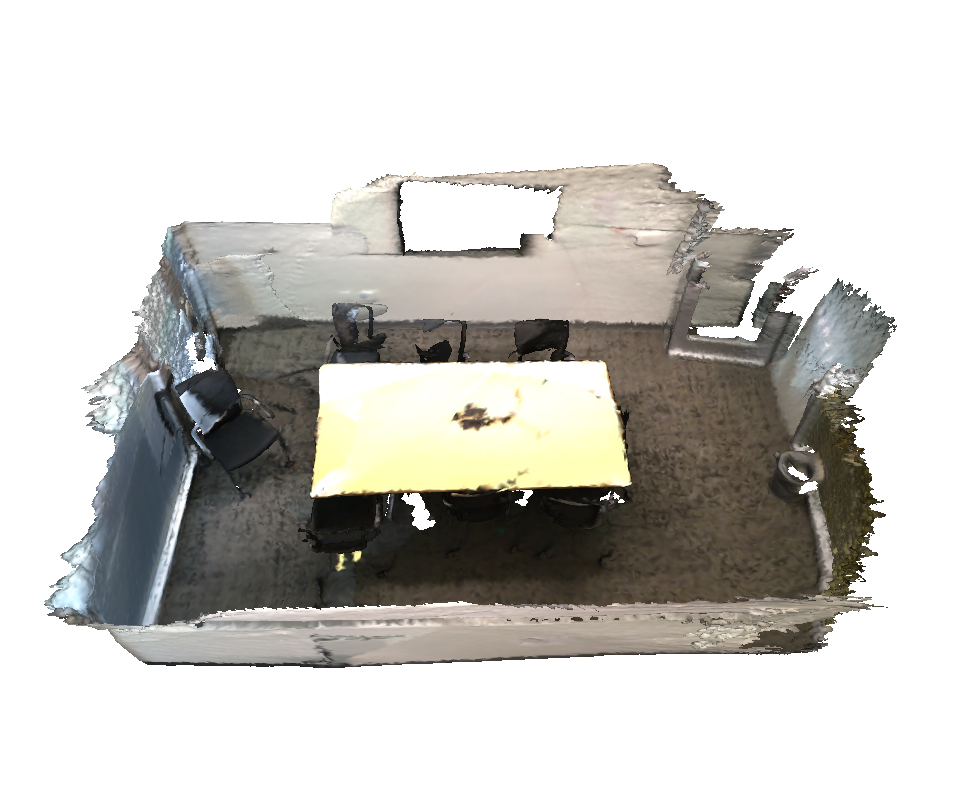}
			\vspace{7pt}
			\includegraphics[width=1.15\linewidth, trim=80 0 0 150,clip]{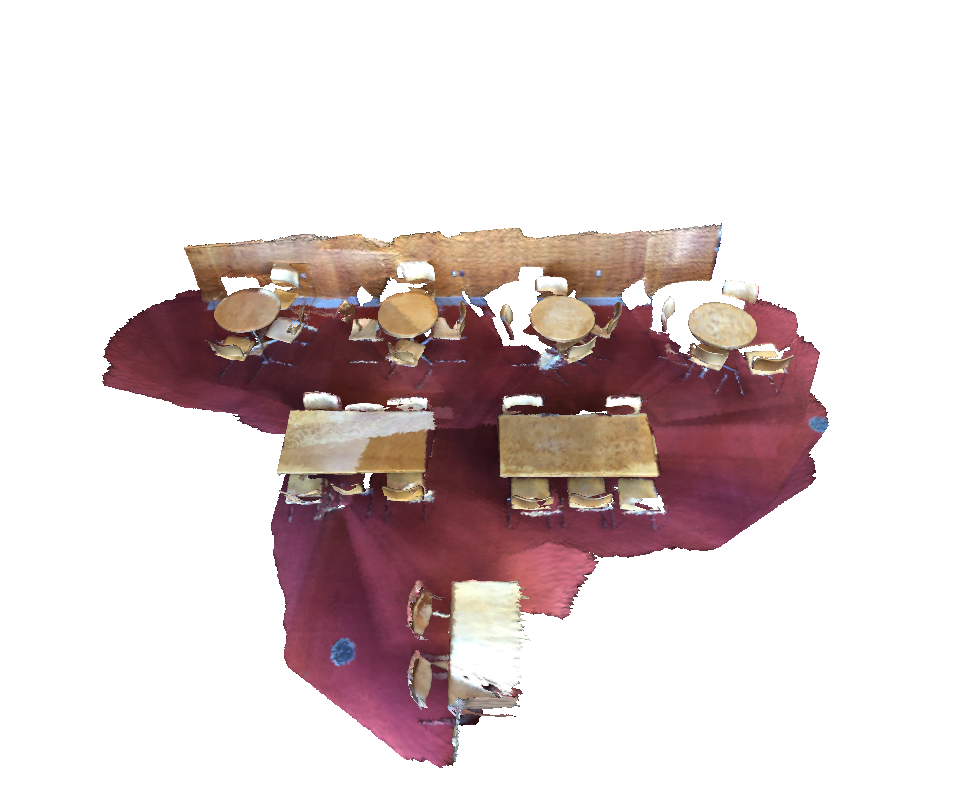}
			\vspace{1pt}
		\end{minipage}
	}
	\subfigure[Seg-level Labels]{
		\begin{minipage}[b]{0.23\linewidth}
			\includegraphics[width=1\linewidth, trim=0 0 0 0,clip]{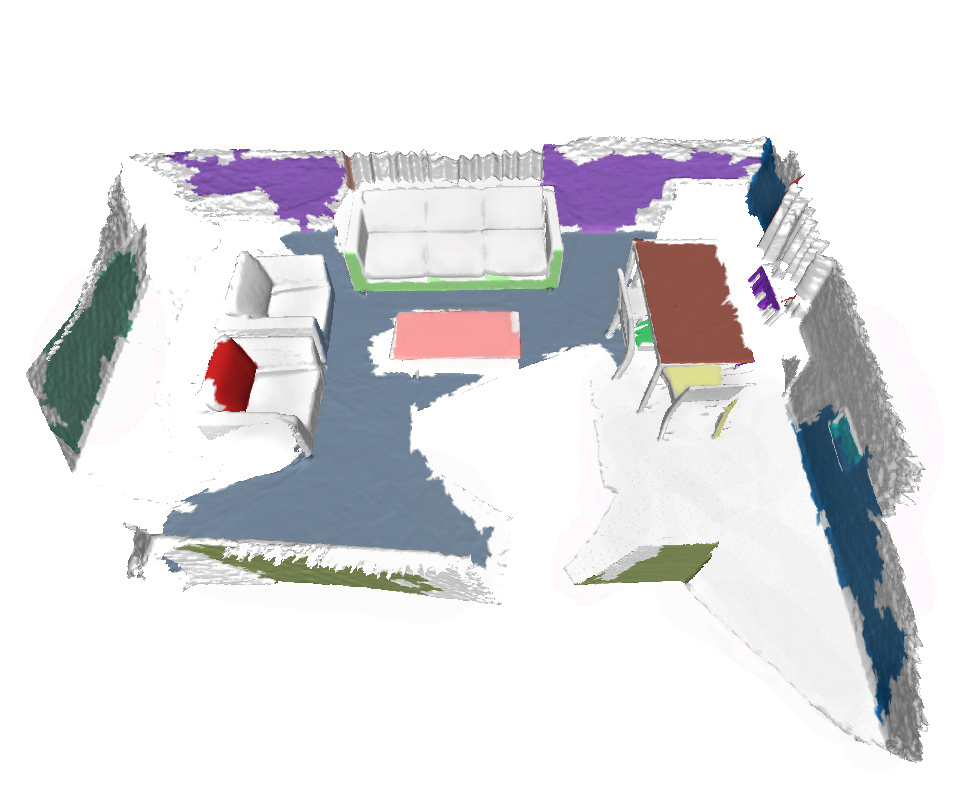}
			\vspace{7pt}
			\includegraphics[width=1\linewidth, trim=0 0 0 0,clip]{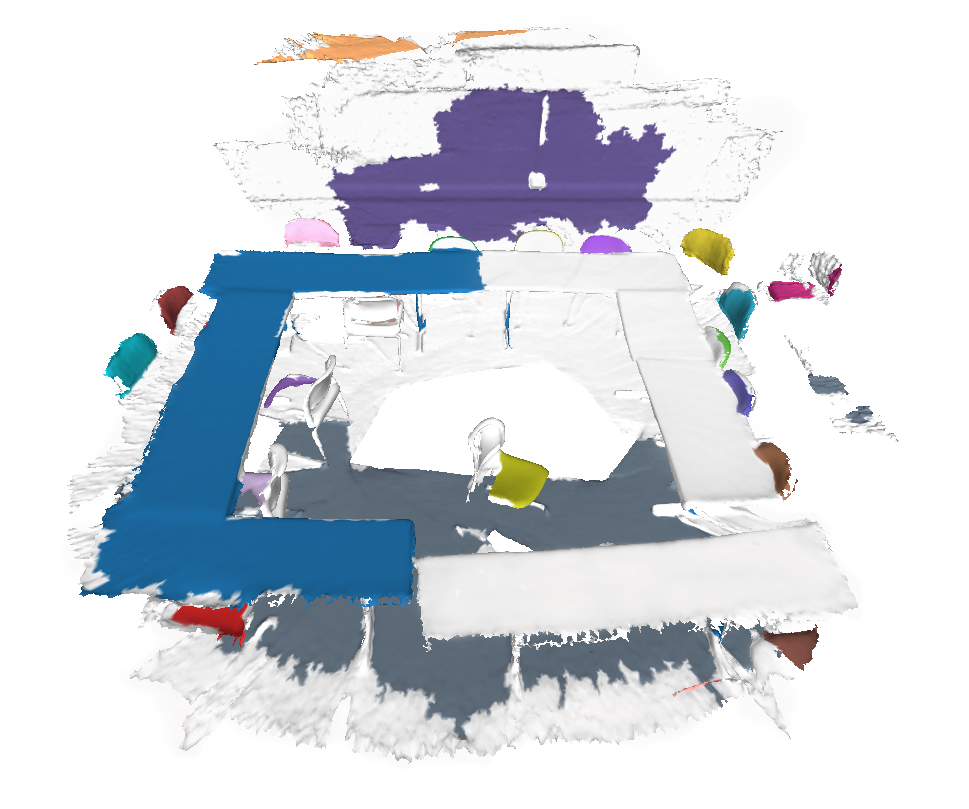}
			\vspace{7pt}
			\includegraphics[width=1\linewidth, trim=0 0 0 0,clip]{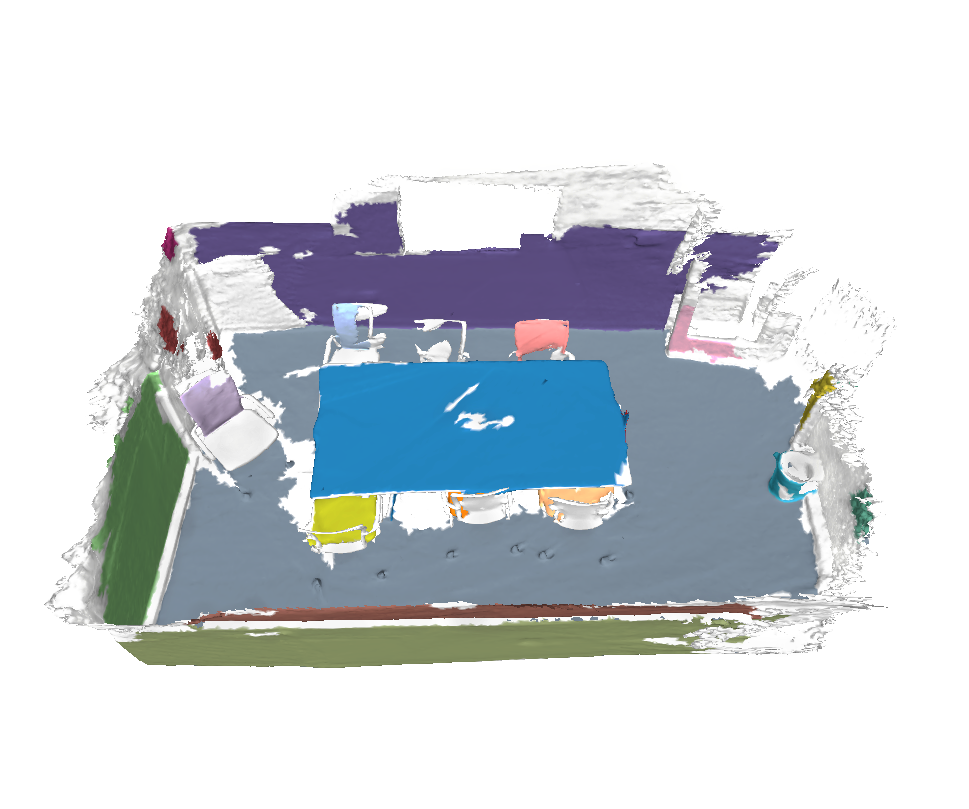}
			\vspace{7pt}
			\includegraphics[width=1.15\linewidth, trim=80 0 0 150,clip]{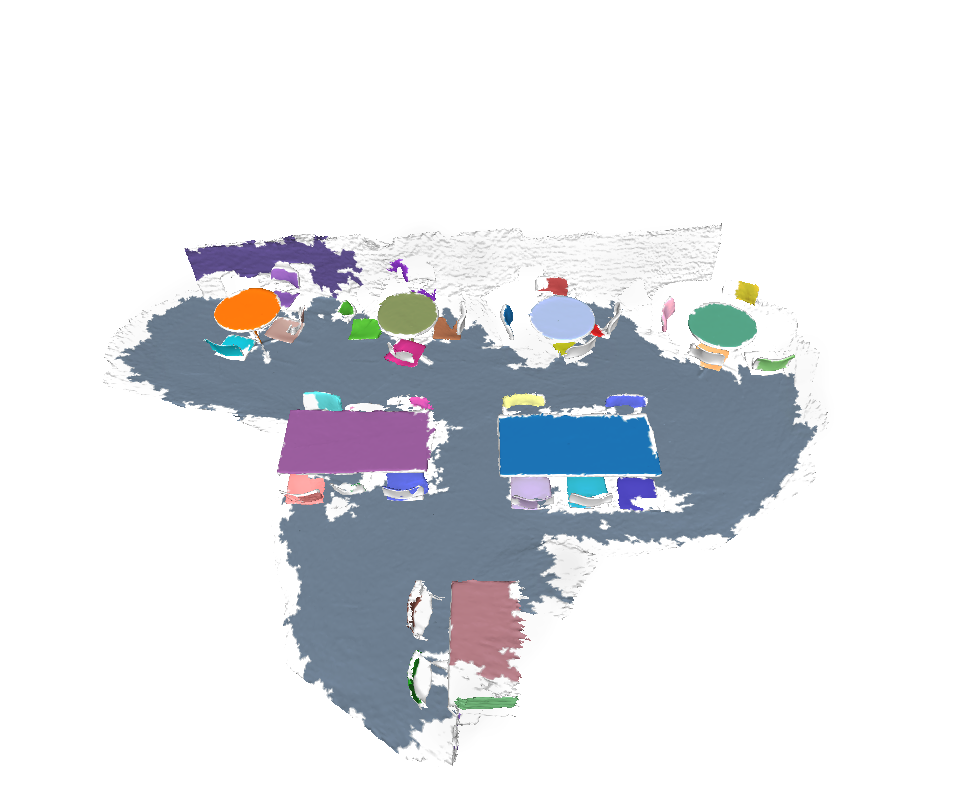}
			\vspace{1pt}
		\end{minipage}
	}
	\subfigure[Pseudo Labels]{
		\begin{minipage}[b]{0.23\linewidth}
			\includegraphics[width=1\linewidth, trim=0 0 0 0,clip]{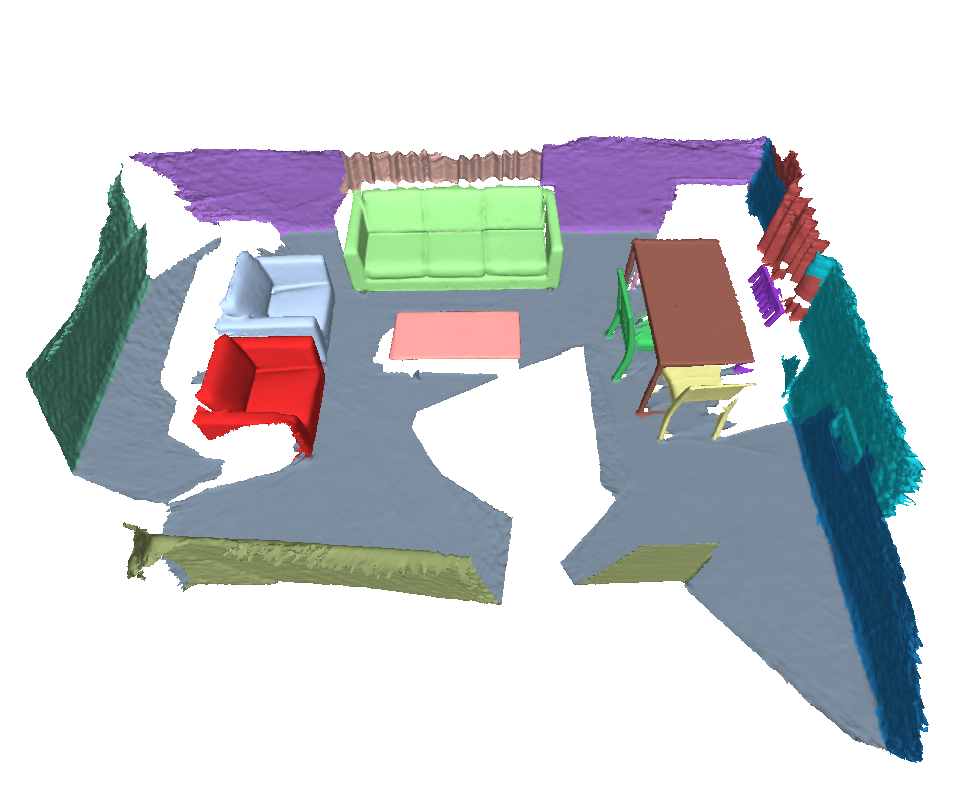}
			\vspace{7pt}
			\includegraphics[width=1\linewidth, trim=0 0 0 0,clip]{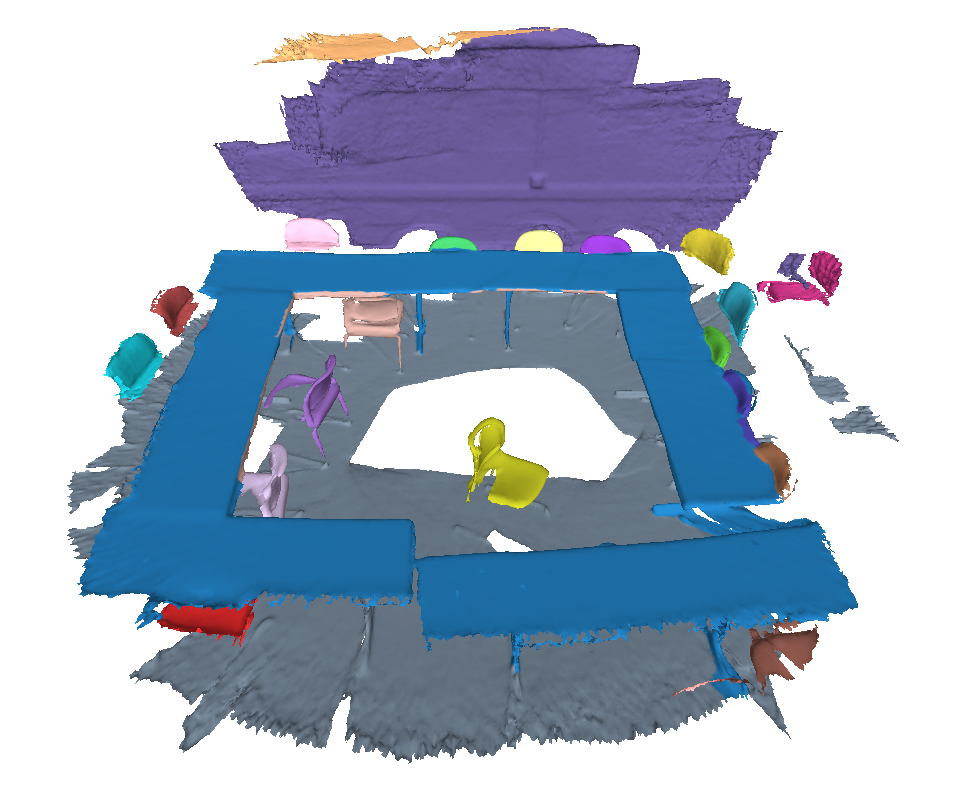}
			\vspace{7pt}
			\includegraphics[width=1\linewidth, trim=0 0 0 0,clip]{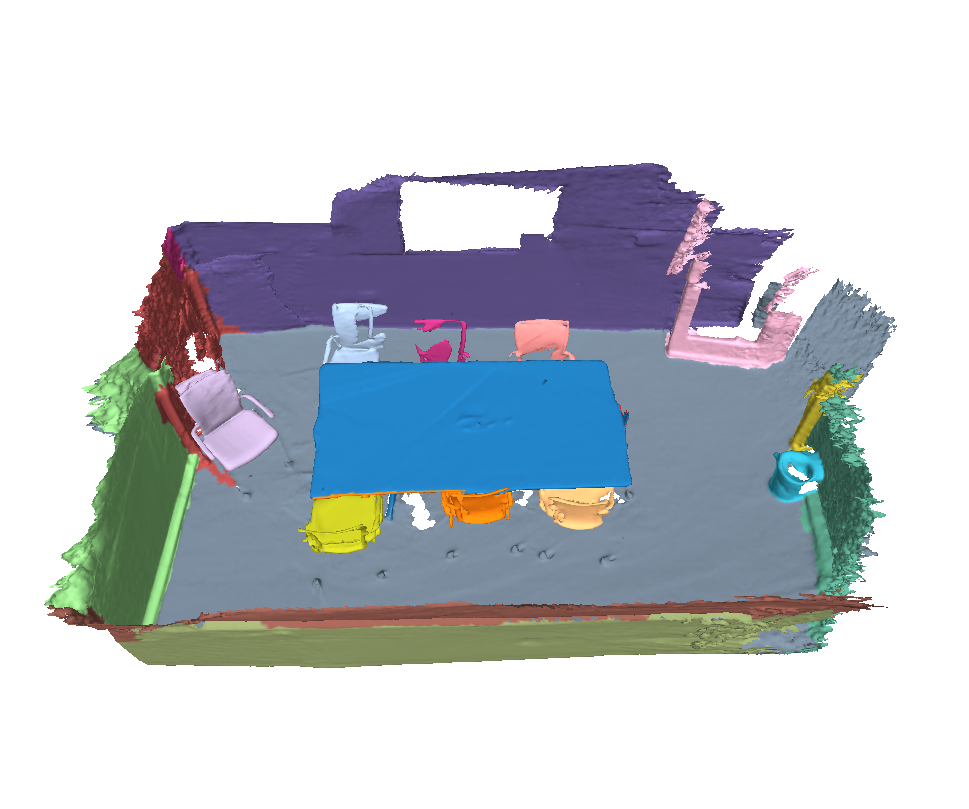}
			\vspace{7pt}
			\includegraphics[width=1.15\linewidth, trim=80 0 0 150,clip]{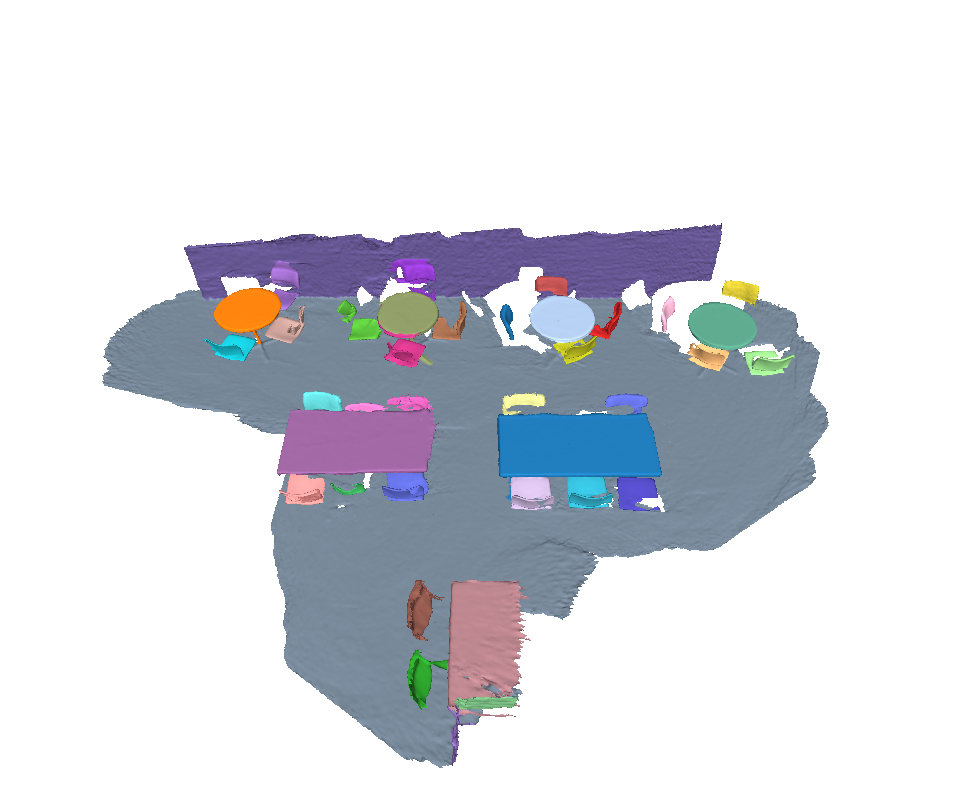}
			\vspace{1pt}
		\end{minipage}
		
	}
	\subfigure[Ground-Truth Labels]{
		\begin{minipage}[b]{0.23\linewidth}
			\includegraphics[width=1\linewidth, trim=0 0 0 0,clip]{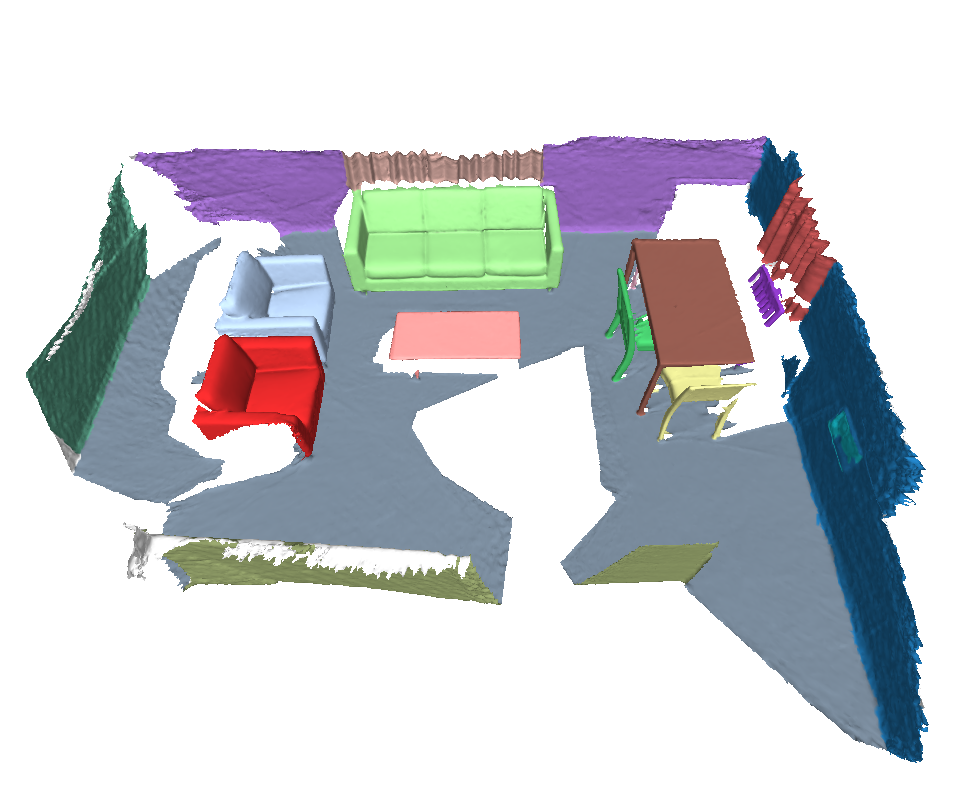}
			\vspace{7pt}
			\includegraphics[width=1\linewidth, trim=0 0 0 0,clip]{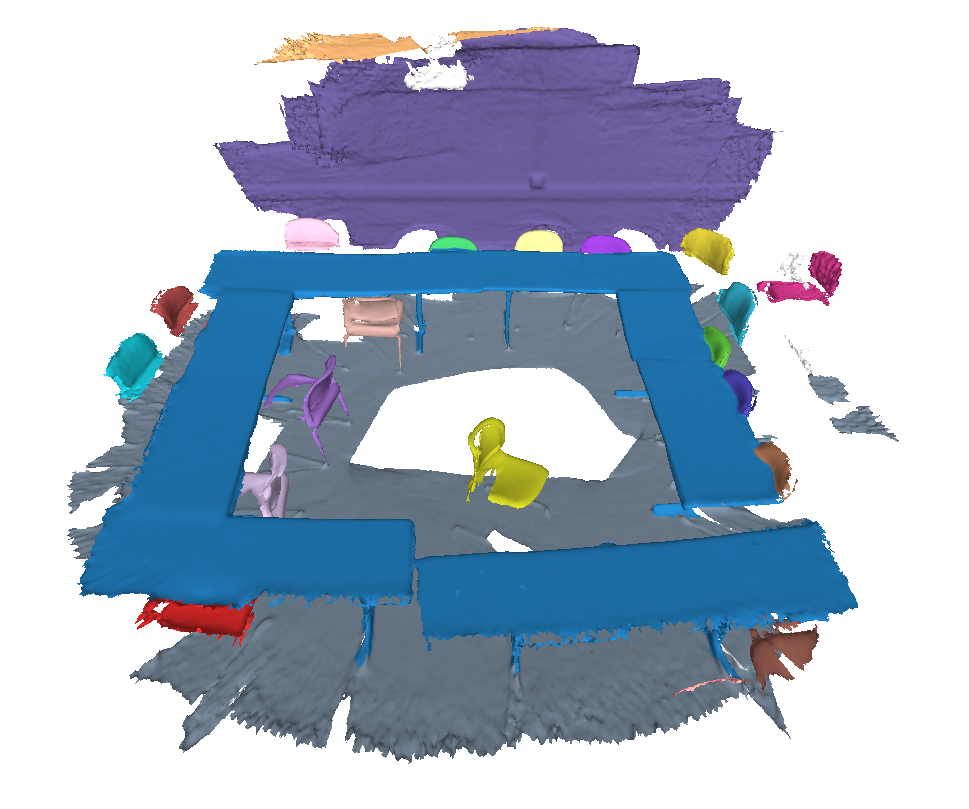}
			\vspace{7pt}
			\includegraphics[width=1\linewidth, trim=0 0 0 0,clip]{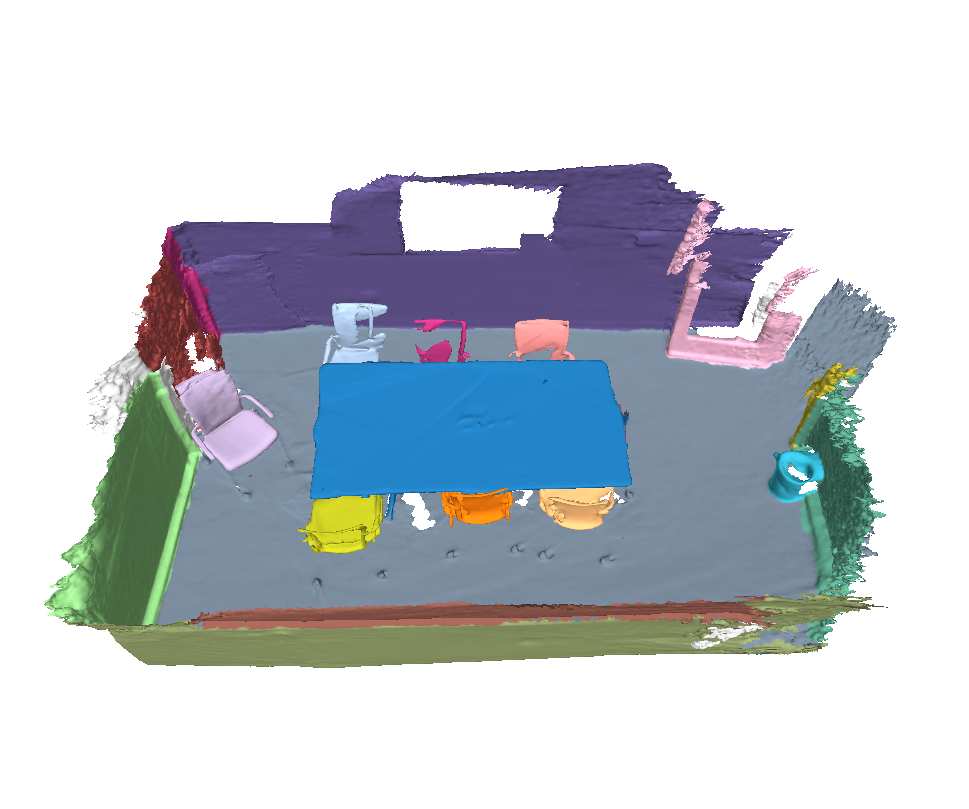}
			\vspace{7pt}
			\includegraphics[width=1.15\linewidth, trim=80 0 0 150,clip]{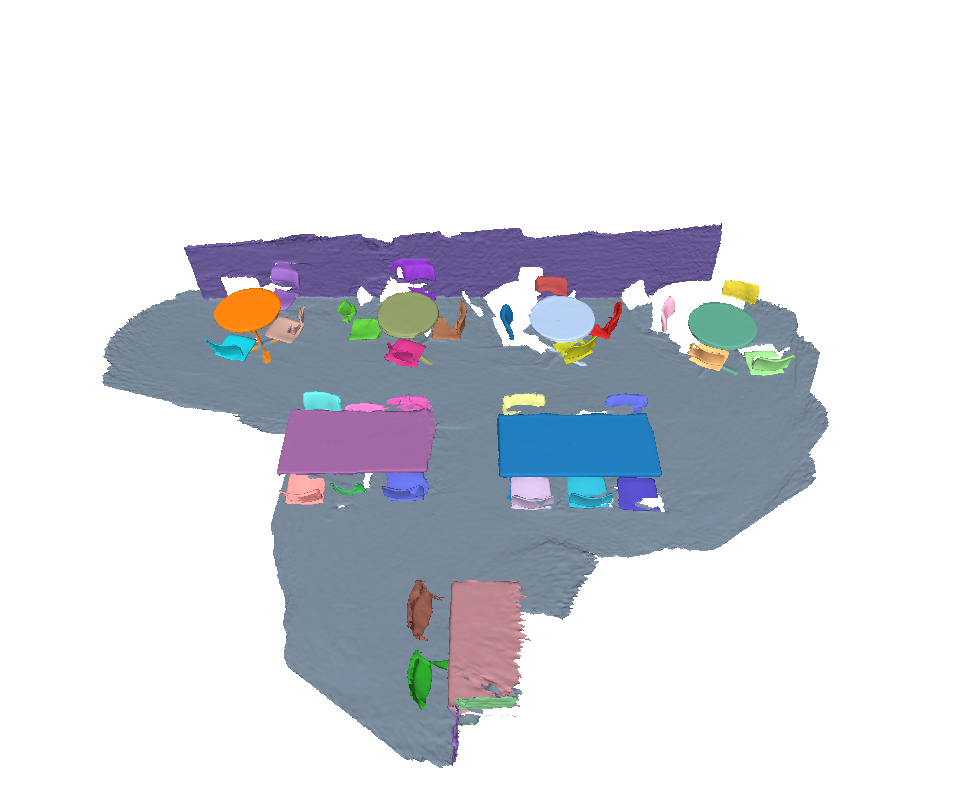}
			\vspace{1pt}
		\end{minipage}
		
	}
	\vspace{7pt}
	\caption{The qualitative visualization results of pseudo label generation. In (b), (c) and (d), various colors indicate different (pseudo) instance labels and points in white are unlabeled. With the benefit of seg-level labels, our SegGroup network can generate qualitative point-level pseudo labels from instance locations. (Best viewed in color.)}
	\label{pseudo}
	\vspace{1pt}
\end{figure*}

\subsection{Implementation Details}
The input of our architecture includes a point cloud scene with RGB colors and a segment graph. Each point in the point cloud scene is 6-dim (XYZ and RGB). The initial node feature $C^1$ in the segment graph is vacant, so the dimension is 0.
The feature extractor in the first semantic layers extracts 64-dim features with one shared multi-layer perceptron (MLP) layer (64), while the extractor in the second layer adopts two shared MLP layers (64, 64) to obtain 64-dim features. After the SegGroup network, the output node feature $C^4$ is 256-dim. Table~\ref{para} lists the details of the dimensions of the features in our architecture.
In all GCNs, we set the output feature dimension the same as the input dimension and $\lambda=1/8$. We set the threshold $e_{\tau}^l$ for clustering as 6 and 2 for the structural grouping layer and the semantic grouping layer, respectively. We have tried many other thresholds (5 and 7 for the structural grouping layer, 1 and 3 for the semantic grouping layer) and found the floating range of the mIoU of generated pseudo labels is within 3\% in semantic mIoU.
This experimental result shows that our network is not very sensitive to the thresholds. 
%As in every layer the node features are concatenated with new segment features, the output feature dimension of the SegGroup network is $128+64+64=256$. 
The classifier obtains class scores with two fully-connected layers. 
We apply dropout with a keep probability of 0.5 in the classifier. 
All layers in the feature extractors and the classifier include LeakyReLU and batch normalization. 
In the network training, we use SGD~\cite{bottou2010large} to optimize the SegGroup network with the learning rate as 0.1 and momentum as 0.9. 
We train the network for 6 epochs in a batch size of 8. 

%In details, there are slight differences among the three feature extractors in the SegGroup network. For the extractor in the structural grouping layer, we first sample 64 points in each segment with farthest point sampling (FPS) and normalize them into a unit sphere. Therefore, the input 6-dim point cloud vectors (XYZ and RGB) only retain geometry and color information which is used to group similarly structured segments into one segment. After extracting local features for each point in EdgeConv, we conduct both a max-pooling and average-pooling on point clouds to obtain 128-dim features. The feature extractors in semantic grouping layers are standard EdgeConv. Besides XYZ and RGB, we also use an additional 3-dim centered coordinates based on its belonged segment to represent each point. 

\section{Experiments}
In this section, we first evaluated the accuracy of the generated pseudo labels compared with ground truth labels. Then, we conducted experiments to show the performance of point cloud instance and semantic segmentation with our pseudo labels. Finally, we compared the annotation efficiency of different label forms with a fixed labeling time. 

We adopted the ScanNet~\cite{dai2017scannet} dataset to conduct our experiment. ScanNet is a widely used large-scale real-world indoor 3D dataset, containing 1,201 training scenes, 312 validation scenes, and 100 hidden test scenes. The dataset has 40 object classes, and the evaluation of semantic segmentation is conducted on 20 classes. For instance segmentation, the wall and floor classes are ignored and only 18 classes are used for evaluation. We annotated seg-level labels in all the 40 classes, where we trained the SegGroup network on these labeled classes. We report the results on 18 classes for instance segmentation tasks and 20 classes for semantic segmentation tasks.

\subsection{Pseudo Label Evaluation}

We first evaluated the generated pseudo labels by instance and semantic IoUs. Table~\ref{sem_iou} shows the class-specific semantic IoU results of the pseudo labels on the ScanNet training set. The semantic IoUs are calculated per semantic class. We also tested the pseudo labels generated in every layer of the SegGroup network. The last row ``SegGroup'' indicates the final output of the network. As the label information is propagated from labeled segments into unlabeled segments with the number of grouping layers increasing, more points in the point cloud scene are annotated with the pseudo labels until all points are annotated.
Experiments show that the generated pseudo labels are very close to the ground truth labels by annotating one point per instance. From the results on different classes, we observe that our method performs significantly better on the classes of \emph{wall} and \emph{floor} which have simple structures as well as \emph{chair}, \emph{sofa}, \emph{table}, \emph{shower}, \emph{curtain} and \emph{toilet} that are easy to identify and separated from a smooth surface. 
Table~\ref{sem_iou} also shows the evaluation results of the pseudo labels generated from scene-level labels and subcloud-level labels by MPRM~\cite{wei2020multi}, where our generated pseudo labels outperform them by a large gap. The mean IoU at Layer 1 of SegGroup is already higher than that of the scene-level labels by MPRM. Considering our method also requires less annotation time per scene compared with the subcloud-level method, experimental results show that our method has a better balance between the annotation time and the quality of the generated pseudo labels.

Besides the quantitative results that compare our generated pseudo labels with the ground truth, we also show qualitative visualizations for a more intuitive illustration in Fig.~\ref{pseudo}. On the ScanNet dataset, there are still a very small number of points remaining unlabeled for point-level annotation, so we can also observe some white details in Fig.~\ref{pseudo}(d). By only annotating one point per instance for its location, we obtain the labels for the corresponding segments with a number of points. In contrast to point-level labels, our labeling method is very cost-effective. With the benefit of seg-level labels, our SegGroup network can generate qualitative point-level pseudo labels from instance locations. From the visualization results, we observe that the pseudo labels match the ground-truth labels in almost all the areas. 

\begin{table}[t]
	\footnotesize
	\caption{Point cloud instance segmentation results (\%) on the ScanNet testing set with different supervisions.}
	\vspace{-15pt}
	\begin{center}
		\begin{tabular}{l|c|c|ccc}
			\toprule
			\hspace{-3pt}Method & \hspace{-4pt}Point Anno.\hspace{-4pt} & \hspace{-5pt}Publication\hspace{-2pt} & AP\hspace{-2pt} & \hspace{-5pt}AP{\tiny 50}\hspace{-5pt} & \hspace{-2pt}AP{\tiny 25}\hspace{-5pt}\\
			\midrule
			\hspace{-3pt}Point-level Supervision:~~ &&&&&\\
			\hspace{-3pt}HAIS~\cite{chen2021hierarchical} & 100\% & ICCV'21 & 45.7\hspace{-2pt} & \hspace{-5pt}69.9\hspace{-5pt} & \hspace{-2pt}80.3\hspace{-5pt}\\
			\hspace{-3pt}SSTNet~\cite{liang2021instance} & 100\% & ICCV'21 & 50.6\hspace{-2pt} & \hspace{-5pt}69.8\hspace{-5pt} & \hspace{-2pt}78.9\hspace{-5pt}\\
			\hspace{-3pt}OccuSeg~\cite{han2020occuseg} & 100\% & CVPR'20 & 48.6\hspace{-2pt} & \hspace{-5pt}67.2\hspace{-5pt} & \hspace{-2pt}78.8\hspace{-5pt}\\
			\hspace{-3pt}PE~\cite{zhang2021point} & 100\% & CVPR'21 & 39.6\hspace{-2pt} & \hspace{-5pt}64.5\hspace{-5pt} & \hspace{-2pt}77.6\hspace{-5pt}\\
			\hspace{-3pt}PointGroup~\cite{jiang2020pointgroup} & 100\% & CVPR'20 & 40.7\hspace{-2pt} & \hspace{-5pt}63.6\hspace{-5pt} & \hspace{-2pt}77.8\hspace{-5pt}\\
			\hspace{-3pt}Dyco3D~\cite{he2021dyco3d} & 100\% & CVPR'21 & 39.5\hspace{-2pt} & \hspace{-5pt}64.1\hspace{-5pt} & \hspace{-2pt}76.1\hspace{-5pt}\\
			\hspace{-3pt}3D-MPA~\cite{engelmann20203d} & 100\% & CVPR'20 & 35.5\hspace{-2pt} & \hspace{-5pt}61.1\hspace{-5pt} & \hspace{-2pt}73.7\hspace{-5pt}\\
			\hspace{-3pt}MTML~\cite{lahoud20193d} & 100\% & ICCV'19 & 28.2\hspace{-2pt} & \hspace{-5pt}54.9\hspace{-5pt} & \hspace{-2pt}73.1\hspace{-5pt}\\
			\hspace{-3pt}3D-BoNet~\cite{yang2019learning} & 100\% & \hspace{-2pt}NeurIPS'19\hspace{-2pt} & 25.3\hspace{-2pt} & \hspace{-5pt}48.8\hspace{-5pt} & \hspace{-2pt}68.7\hspace{-5pt}\\
			\hspace{-3pt}3D-SIS~\cite{hou20193d} & 100\% & CVPR'19 & 16.1\hspace{-2pt} & \hspace{-5pt}38.2\hspace{-5pt} & \hspace{-2pt}55.8\hspace{-5pt}\\
			\hspace{-3pt}GSPN~\cite{yi2019gspn} & 100\% & CVPR'19 & 15.8\hspace{-2pt} & \hspace{-5pt}30.6\hspace{-5pt} & \hspace{-2pt}54.4\hspace{-5pt}\\
			\hspace{-3pt}SGPN~\cite{wang2018sgpn} & 100\% & CVPR'18 & 4.9\hspace{-2pt} & \hspace{-5pt}14.3\hspace{-5pt} & \hspace{-2pt}39.0\hspace{-5pt}\\
			\midrule
			%		\hspace{-3pt}One Thing One Click: &&&&&\\
			%		\hspace{-3pt}OTOC (PointGroup)~\cite{liu2021one} & 0.022\% & CVPR'21 & 2.4\hspace{-2pt} & \hspace{-5pt}5.6\hspace{-5pt} & \hspace{-2pt}13.3\hspace{-5pt}\\
			%		\midrule
			\hspace{-3pt}Init+Act. Point Supervision:\hspace{-5pt} &&&&&\\
			\hspace{-3pt}CSC-20 (PointGroup)~\cite{hou2021exploring} & 0.014\% & CVPR'21 & 15.9\hspace{-2pt} & \hspace{-5pt}28.9\hspace{-5pt} & \hspace{-2pt}49.6\hspace{-5pt}\\
			\hspace{-3pt}CSC-50 (PointGroup)~\cite{hou2021exploring} & 0.034\% & CVPR'21 & 22.9\hspace{-2pt} & \hspace{-5pt}41.4\hspace{-5pt} & \hspace{-2pt}62.0\hspace{-5pt}\\
			\midrule
			\hspace{-3pt}Seg-level Supervision: &&&&&\\
			\hspace{-3pt}SegGroup (PointGroup) & 0.028\% & - & 24.6\hspace{-2pt} & \hspace{-5pt}44.5\hspace{-5pt} & \hspace{-2pt}63.7\hspace{-5pt}\\
			\bottomrule
		\end{tabular}
	\end{center}
	\label{ins_test}
	%	\vspace{10pt}
\end{table}

\begin{table}[t]
	\footnotesize
	\caption{Point cloud instance segmentation results (\%) on the ScanNet validation set with a fixed labeling time budget on the training set.}
	\vspace{-7pt}
	\begin{center}
		\begin{tabular}{l|c|ccc}
			\toprule
			Method & Scenes & ~AP & \hspace{-3pt}AP{\tiny 50}\hspace{-3pt} & AP{\tiny 25}\\
			\midrule
			Point-level Supervision: &&&&\\
			PointGroup~\cite{jiang2020pointgroup} & 104 & ~9.4 & \hspace{-3pt}19.9\hspace{-3pt} & 36.4\\
			\midrule
			Seg-level Supervision: &&&&\\
			SegGroup (PointGroup) & 1201 & ~23.4 & \hspace{-3pt}43.4\hspace{-3pt} & 62.9 \\
			\bottomrule
		\end{tabular}
	\end{center}
	\label{ins_time}
	\vspace{-10pt}
\end{table}

\subsection{Point Cloud Instance Segmentation}
\label{ins_seg}
For the evaluation of the point cloud instance segmentation task, we employed PointGroup~\cite{jiang2020pointgroup} to train a point cloud instance segmentation model based on the generated point-level pseudo labels. Table~\ref{ins_test} shows the recent point-level and weakly-supervised point cloud instance segmentation results on the ScanNet testing set. 
The Point Anno. entry denotes the percentage of manually annotated point labels in the total points of the point cloud scene. Because the numbers of segment labels and point labels are the same in our method, we consider a segment label as a point label when computing the percentage of overall annotated data.
%We use mean average precision (mAP) for the evaluation metric. 
AP averages the scores with IoU (Intersection over Union) threshold set from 50\% to 95\% with a step size of 5\%, while AP{\scriptsize 25} and AP{\scriptsize 50} denote the AP scores with IoU threshold set as 25\% and 50\% respectively. 
We use CSC-20 and CSC-50 to denote the Contrastive Scene Contexts (CSC)~\cite{hou2021exploring} method that is annotated with 20 and 50 points per scene. 
%The result of OTOC~\cite{liu2021one} is implemented on our own to validate that it cannot be directly used for the instance segmentation task.
We observe that our seg-level supervised method achieves competitive results with the recent strong supervised methods. However, the seg-level labels only require 1.93 minutes to click on 0.028\% of points for one scene in ScanNet on average, while the strong point-level labels need 22.3 minutes. Compared with the CSC~\cite{hou2021exploring} method which proposes an active labeling strategy (denoted as init+act.) to annotate points, our annotations are per instance and indicate the locations. In seg-level labels, we annotate 41.2 points on average for each 3D scene. The results show that our seg-level labels obtain much better performance than init+act. point annotations for the instance segmentation task. 
%Our method also shows great superiority against OTOC~\cite{liu2021one} on the instance segmentation task compared. 

Besides the experiments above, we also compared the efficiency of different labeling strategies given a fixed annotation time budget on the training set. Table~\ref{ins_time} shows point cloud instance segmentation results under different supervisions on the ScanNet validation set with different numbers of training scenes. Given the labeling time for the whole training set (1201 scenes) with seg-level labels, only 104 scenes can be annotated with point-level labels. The results show that our seg-level labels obtain much better performance than point-level labels given the same annotation budget for the instance segmentation task.

\begin{table}[t]
	\footnotesize
	\caption{Point cloud semantic segmentation results (\%) on the ScanNet testing set compared with different supervisions. }
	\begin{center}
		\begin{tabular}{l|c|c|c}
			\toprule
			Method & Point Anno. & Publication & mIoU\\
			\midrule
			Point-level Supervision: &&&\\
			Mix3D~\cite{nekrasov2021mix3d} & 100\% & 3DV'21 & 78.1 \\
			OccuSeg~\cite{han2020occuseg} & 100\% & CVPR'20 & 76.4 \\
			VMNet~\cite{hu2021vmnet} & 100\% & ICCV'21 & 74.6 \\
			Virtual MVFusion~\cite{kundu2020virtual} & 100\% & ECCV'20 & 74.6 \\
			MinkowskiNet~\cite{choy20194d} & 100\% & CVPR'19 & 73.4 \\
			SparseConvNet~\cite{graham20183d} & 100\% & CVPR'18 & 72.5 \\
			JSENet~\cite{hu2020jsenet} & 100\% & ECCV'20 & 69.9 \\
			FusionNet~\cite{zhang2020deep} & 100\% & ECCV'20 & 68.8 \\
			KPConv~\cite{thomas2019kpconv} & 100\% & ICCV'19 & 68.4 \\
			DCM-Net~\cite{schult2020dualconvmesh} & 100\% & CVPR'20 & 65.8 \\
			RandLA-Net~\cite{hu2020randla} & 100\% & CVPR'20 & 64.5 \\
			FusionAwareConv~\cite{zhang2020fusion} & 100\% & CVPR'20 & 63.0 \\
			HPEIN~\cite{jiang2019hierarchical} & 100\% & ICCV'19 & 61.8 \\
			SegGCN~\cite{lei2020seggcn} & 100\% & CVPR'20 & 58.9 \\
			TextureNet~\cite{huang2019texturenet} & 100\% & CVPR'19 & 56.6 \\
			3DMV~\cite{dai20183dmv} & 100\% & ECCV'18 & 48.8 \\
			PointCNN~\cite{li2018pointcnn} & 100\% & NeurIPS'18 & 45.8 \\
			%			FCPN~\cite{rethage2018fully} & ECCV'18 & 44.2 \\
			PointNet++~\cite{qi2017pointnet++} & 100\% & NeurIPS'17 & 33.9 \\
			ScanNet~\cite{dai2017scannet} & 100\% & CVPR'17 & 30.6 \\
			\midrule
			Subcloud-level Supervision: &&&\\
			MPRM (KPConv)~\cite{wei2020multi} & Subcloud & CVPR'20 & 41.1 \\
			\midrule
			Rand Point Supervision: &&&\\
			PSD (RandLA-Net)~\cite{zhang2021perturbed} & 1\% & ICCV'21 & 54.7 \\
			SQN (RandLA-Net)~\cite{hu2021sqn} & 0.1\% & ECCV'22 & 53.5 \\
			\midrule
			One Thing One Click: &&&\\
			OTOC (SparseConvNet)~\cite{liu2021one} & 0.022\% & CVPR'21 & 69.1 \\
			\midrule
			Init+Act. Point Supervision: &&&\\
			CSC-20 (MinkowskiNet)~\cite{hou2021exploring} & 0.014\% & CVPR'21 & 53.1\\
			OTOC-20 (SparseConvNet)~\cite{liu2021one} & 0.014\% & CVPR'21 & 59.4 \\
			CSC-50 (MinkowskiNet)~\cite{hou2021exploring} & 0.034\% & CVPR'21 & 61.2\\
			OTOC-50 (SparseConvNet)~\cite{liu2021one} & 0.034\% & CVPR'21 & 64.2 \\
			\midrule
			Seg-level Supervision: &&&\\
			SegGroup (KPConv) & 0.028\% & - & 61.1 \\
			SegGroup (MinkowskiNet) & 0.028\% & - & 62.7 \\
			\bottomrule
		\end{tabular}
	\end{center}
	\label{sem_test}
	\vspace{-5pt}
\end{table}

\subsection{Point Cloud Semantic Segmentation}
For the point cloud semantic segmentation task, we trained MinkowskiNet~\cite{choy20194d} and KPConv~\cite{thomas2019kpconv} with the generated pseudo labels for evaluation. %\footnote{We discover that the semantic segmentation task is more sensitive to incorrect pseudo labels than no label. To generate pseudo labels after the training process of the SegGroup, we choose to propagate label information from labeled segments only into the segments where the confidence is relatively high. The partially generated pseudo labels are adopted to train the point-level supervised point cloud segmentation model.}
Table~\ref{sem_test} shows the results of both the recent point-level and other weak supervision methods\footnote{If an instance is composed of some disconnected portions, we allow the annotator to label one segment on each of them. Therefore, the point annotation rate of SegGroup is slightly higher than OTOC~\cite{liu2021one}.} on the ScanNet testing set. Following the definitions in Table~\ref{ins_test}, we use OTOC-20 and OTOC-50 to denote the OTOC~\cite{liu2021one} method that is annotated with 20 and 50 points per scene.
We use mIoU (mean Intersection over Union) as the evaluation metric. 
%We observe that our SegGroup significantly outperforms the subcloud-level supervised method with the same KPConv network for semantic segmentation, which shows the effectiveness of the seg-level labels. While subcloud-level labels need 3 minutes to annotate a scene, our seg-level labels only require about 1 minute. Therefore, our SegGroup achieves better semantic segmentation performance with fewer annotation costs compared with the subcloud-level labels. 
The results show that we achieve comparable results with the strong point-level supervised methods. Considering the huge time cost of the point-level labels, our seg-level supervised method is very competitive. Moreover, we outperform most of the weakly-supervised methods in a comparable point annotation rate. Considering the annotation time of the subcloud-level labels in MPRM~\cite{wei2020multi} is about 3 minutes per scene, we also surpass MPRM on the semantic segmentation performance with a less time budget of 1.93 minutes per scene. Experimental results show that our seg-level labels provide a nice tradeoff between annotation time and segmentation accuracy, which shows that precise instance locations are crucial for 3D scene understanding. %Moreover, annotating instance location as seg-level labels can be a low-cost but high-yield labeling manner for 3D instance and semantic segmentation.

\begin{table*}[t]
	\footnotesize
	\caption{The class-specific semantic IoUs (\%) of point cloud semantic segmentation results on the ScanNet validation set  compared with the fully-supervised method. }
	\vspace{-7pt}
	{\scriptsize
		\begin{center}
			\begin{tabular}{l|c|m{0.2cm}<{\centering}m{0.2cm}<{\centering}m{0.2cm}<{\centering}m{0.2cm}<{\centering}m{0.2cm}<{\centering}m{0.2cm}<{\centering}m{0.2cm}<{\centering}m{0.2cm}<{\centering}m{0.2cm}<{\centering}m{0.2cm}<{\centering}m{0.2cm}<{\centering}m{0.2cm}<{\centering}m{0.2cm}<{\centering}m{0.2cm}<{\centering}m{0.2cm}<{\centering}m{0.2cm}<{\centering}m{0.2cm}<{\centering}m{0.2cm}<{\centering}m{0.2cm}<{\centering}m{0.4cm}<{\centering}|m{0.4cm}<{\centering}}
				\toprule
				Manner & \hspace{-2pt}Point Anno.\hspace{-2pt} & ~\,\rotatebox{90}{Wall} & ~\,\rotatebox{90}{Floor} & ~\,\rotatebox{90}{Cab.} & ~\,\rotatebox{90}{Bed} & ~\,\rotatebox{90}{Chair} & ~\,\rotatebox{90}{Sofa} & ~\,\rotatebox{90}{Table} & ~\,\rotatebox{90}{Door} & ~\,\rotatebox{90}{Wind.} & ~\,\rotatebox{90}{Bshf.} & ~\,\rotatebox{90}{Pic.} & ~\,\rotatebox{90}{Cntr.} & ~\,\rotatebox{90}{Desk} & ~\,\rotatebox{90}{Curt.} & ~\,\rotatebox{90}{Fridg.} & ~\,\rotatebox{90}{Shwr.} & ~\,\rotatebox{90}{Toil.} & ~\,\rotatebox{90}{Sink} & ~\,\rotatebox{90}{Bath.} & \rotatebox{90}{Ofurn.}& \rotatebox{90}{Mean} \\
				\midrule
				Point-level Supervision: &&&&&&&&&&&&&&&&&&&&&\\
				KPConv~\cite{thomas2019kpconv} & 100\% & 82.1 & 94.4 & 64.4 & 79.5 & 89.2 & 77.7 & 72.3 & 59.1 & 58.0 & 78.7 & 28.5 & 60.6 & 62.0 & 70.6 & 50.8 & 51.1 & 91.7 & 62.6 & 85.9 & 54.0 & 68.7 \\
				MinkowskiNet~\cite{choy20194d} & 100\% & 84.8 & 94.9 & 65.6 & 80.4 & 90.5 & 85.4 & 74.6 & 64.0 & 62.1 & 80.8 & 32.4 & 61.6 & 63.8 & 77.6 & 53.8 & 69.9 & 91.9 & 69.6 & 87.3 & 59.3 & 72.5 \\
				\midrule
				Seg-level Supervision: &&&&&&&&&&&&&&&&&&&&&\\
				SegGroup (KPConv) & 0.028\% & 78.1 & 92.9 & 52.6 & 69.3 & 82.8 & 69.4 & 66.7 & 55.9 & 51.1 & 73.5 & 29.4 & 56.0 & 50.3 & 61.0 & 37.6 & 56.3 & 83.1 & 57.0 & 77.8 & 46.5 & 62.4 \\
				SegGroup (MinkowskiNet) & 0.028\% & 78.8 & 93.3 & 53.7 & 73.3 & 83.5 & 74.4 & 67.2 & 55.3 & 49.6 & 72.8 & 27.2 & 57.3 & 54.3 & 63.9 & 43.8 & 66.1 & 87.1 & 58.8 & 80.2 & 49.1 & 64.5 \\
				\bottomrule
			\end{tabular}
		\end{center}
	}
	\label{sem_iou_val}
\end{table*}

\begin{table}[t]
	\footnotesize
	\caption{Point cloud semantic segmentation results (\%) on the ScanNet validation set with a fixed labeling time budget on the training set.}
	\begin{center}
		\begin{tabular}{l|c|c}
			\toprule
			Method & Scenes & mIoU\\
			\midrule
			Point-level Supervision: &&\\
			KPConv~\cite{thomas2019kpconv} & 104 & 53.2 \\
			MinkowskiNet~\cite{choy20194d} & 104 & 53.0 \\
			\midrule
			Seg-level Supervision: &&\\
			SegGroup (KPConv) & 1201 & 62.4 \\
			SegGroup (MinkowskiNet) & 1201 & 64.5 \\
			\bottomrule
		\end{tabular}
	\end{center}
	\label{sem_time}
	%	\vspace{-10pt}
\end{table}

Although the performance of OTOC~\cite{liu2021one} on semantic segmentation is better than ours, OTOC can not be directly adopted for instance segmentation. This limitation is determined by its objective function (energy function) for pseudo label generation, where both the unary term and the pairwise term only consider semantic labels. The energy function is built upon the point-level semantic prediction outputs of the semantic segmentation model. The final semantic pseudo labels are obtained by minimizing the energy function.
%In existing point cloud instance segmentation methods, the instance labels are either aggregated from point-level semantic predictions or generated from the 3D bounding box detection results. For each predicted instance, the instance segmentation model cannot output the instance prediction scores for all instance IDs in the scene.
%	%For point cloud instance segmentation, the instance labels are either aggregated from point-level semantic predictions or generated from the 3D bounding box detection results. 
%	For each predicted instance, the instance segmentation model cannot output the instance prediction scores for all instance IDs in the scene. Therefore, the energy function can only be adopted for semantic labels. 
%due to the semantic nature of its energy function. 
%However, an important emphasis of our method is instance segmentation for point cloud scenes with limited annotations. 
%To validate the superiority of our method in instance segmentation, in Table~\ref{ins_test} we use the original weak instance labels and the generated pseudo semantic labels of OTOC to learn instance segmentation with an instance segmentation model PointGroup~\cite{jiang2020pointgroup}. Experimental results show the superiority of our SegGroup. Our method also has the advantage of fewer computational costs. 
%The parameter number of OTOC is very large, since it adopts two SparseConvNet~\cite{graham20183d} networks separately for the semantic segmentation model and the relation network. 
We also find that our method uses much fewer network parameters than OTOC. When we remove the semantic segmentation model and remain the backbone for pseudo label generation, the parameter numbers are $30.11{\rm M}$ v.s. $0.15{\rm M}$ for OTOC and our SegGroup. 

We presented the semantic IoU per class results after the classical learning process in Table~\ref{sem_iou_val}. For the point-level and seg-level supervised methods, we observe that the rankings of class-specific IoUs from the highest to the lowest are almost the same.  This phenomenon shows that our seg-level supervised method has similar behavior to the point-level supervised method. When comparing Table~\ref{sem_iou_val} with Table~\ref{sem_iou}, the experimental results do not show explicit relation between the rankings of the class-specific IoUs of pseudo labels on the training set in Table~\ref{sem_iou} and the predicted labels on the validation set in Table~\ref{sem_iou_val}. We think the rankings of class-specific semantic IoUs of both point-level and seg-level supervised methods in Table~\ref{sem_iou_val} are determined by the difficulty of the classes themselves in the training process. 

We also compared the efficiency of different labeling strategies on point cloud semantic segmentation following the settings in Sec.~\ref{ins_seg}. Table~\ref{sem_time} shows the point cloud semantic segmentation results under different supervisions on the ScanNet validation set with a fixed annotation budget on the training set. Given the labeling time for the whole training set with seg-level labels, only 104 scenes can be annotated with point-level labels. 
%Compared with the recent method CSC~\cite{hou2021exploring}, we achieve very competitive performance by using less point annotations (41.2 vs. 50) with a more efficient backbone (KPConv~\cite{thomas2019kpconv} vs. MinkowskiNet~\cite{choy20194d}). %The MinkowskiNet adopted by CSC~\cite{hou2021exploring} can achieve 72.2\% in the ScanNet validation set with strong point-level labels, while the KPConv can only achieve 69.2\%.
Our SegGroup outperforms the fully supervised method under the same annotation cost.
%Nevertheless, our SegGroup still outperforms all other methods under the same annotation cost. 
Combining the results in both Table~\ref{ins_time} and Table \ref{sem_time}, our seg-level labels provide a better tradeoff between annotation time and segmentation accuracy compared with point-level methods, which shows that precise instance locations are crucial for 3D scene understanding. Annotating instance location as seg-level labels can be a low-cost but high-yield labeling manner for 3D instance and semantic segmentation.

\begin{table}[t]
	\footnotesize
	\caption{Point cloud instance and semantic segmentation results (\%) on the ScanNet validation set supervised by seg-level labels obtained with different annotation manners.}
	\begin{center}
		\begin{tabular}{l|ccc|c}
			\toprule
			\multirow{2}{*}{Manner} & \multicolumn{3}{c|}{Ins Seg} & Sem Seg \\
			& AP & AP{\tiny 50} & AP{\tiny 25} & mIoU\\
			\midrule
			Fully-Sup Baseline & 34.8 & 56.9 & 71.3 & 68.7 \\
			\midrule
			Mechanical:&&&&\\
			Top-1 Segment & 24.8 & 46.2 & 64.6 & 63.6 \\
			Top-2 Segment & 25.0 & 43.9 & 62.6 & 64.2 \\
			Top-3 Segment & 24.2 & 43.0 & 62.4 & 64.2 \\
			Rand Segment & 20.0 & 36.9 & 57.8 & 46.7 \\
			\midrule
			Manual: &&&&\\
			Top-1 Segment & 23.4 & 43.4 & 62.9 & 62.4 \\
			\bottomrule
		\end{tabular}
	\end{center}
	\label{label_type}
	%	\vspace{-10pt}
\end{table}

\subsection{Ablation Study}
Before manual labeling started, we first generated various types of weak labels from ground-truth strong labels and conducted experiments to compare their performance. 
We employed PointGroup~\cite{jiang2020pointgroup} and KPConv~\cite{thomas2019kpconv} to train a point cloud instance segmentation model and a point cloud semantic segmentation model based on the generated point-level pseudo labels from our SegGroup. 
The Mechanical part of Table~\ref{label_type} shows the point cloud instance and semantic segmentation results under various annotation manners on the ScanNet validation set. For mechanical seg-level labels generated from Top-$N$ segments, we choose to annotate one point randomly in the range of Top-$N$ largest segments for each instance. The results show that larger segments can yield better performance. We follow these experimental results to design our manual annotation rule as annotating on the largest segment of an instance as the most representative segment.

We also compared the annotation quality between manual labeling by annotators and mechanical labeling from the ground-truth point-level labels in Table~\ref{label_type}. 
%Since in practice the ground-truth labels are not available, the mechanical labeling only serves as a comparison. 
Compared to mechanical annotations, the results of manual annotations are slightly worse but comparable. Therefore, our manual labeling strategy for seg-level labels is practical for future applications. Moreover, we find that for mechanical annotations the results of Top-3 are better than Top-1 on semantic segmentation, while worse on instance segmentation. Top-3 annotations can increase variety in seg-level labels across instances, while lower the labeling quality of each instance. The results show that semantic segmentation supervision benefits more from labeling variety across instances, while instance segmentation is more sensitive to instance-specific labeling quality.

\begin{table}[t]
	\footnotesize
	\caption{Point cloud instance and semantic segmentation results (\%) on the ScanNet validation set supervised by seg-level labels obtained with different annotation numbers per instance.}
	\begin{center}
		\begin{tabular}{l|ccc|c}
			\toprule
			\multirow{2}{*}{Manner} & \multicolumn{3}{c|}{Ins Seg} & Sem Seg \\
			& AP & AP{\tiny 50} & AP{\tiny 25} & mIoU \\
			\midrule
			Fully-Sup Baseline & 34.8 & 56.9 & 71.3 & 68.7 \\
			\midrule
			Mechanical:&&&&\\
			One Segment & 24.8 & 46.2 & 64.6 & 63.6 \\
			Two Segments & 28.1 & 49.0 & 66.3 & 66.3 \\
			Three Segments & 29.1 & 48.7 & 67.1 & 67.4 \\
			Five Segments & 30.2 & 51.7 & 68.3 & 68.4 \\
			Seven Segments & 31.4 & 52.5 & 68.6 & 68.5 \\
			Ten Segments & 33.1 & 55.3 & 69.3 & 68.7 \\
			\bottomrule
		\end{tabular}
	\end{center}
	\label{anno_num}
	%	\vspace{-10pt}
\end{table}

\begin{figure}[t]
	\begin{center}
		\includegraphics[width=0.8\linewidth]{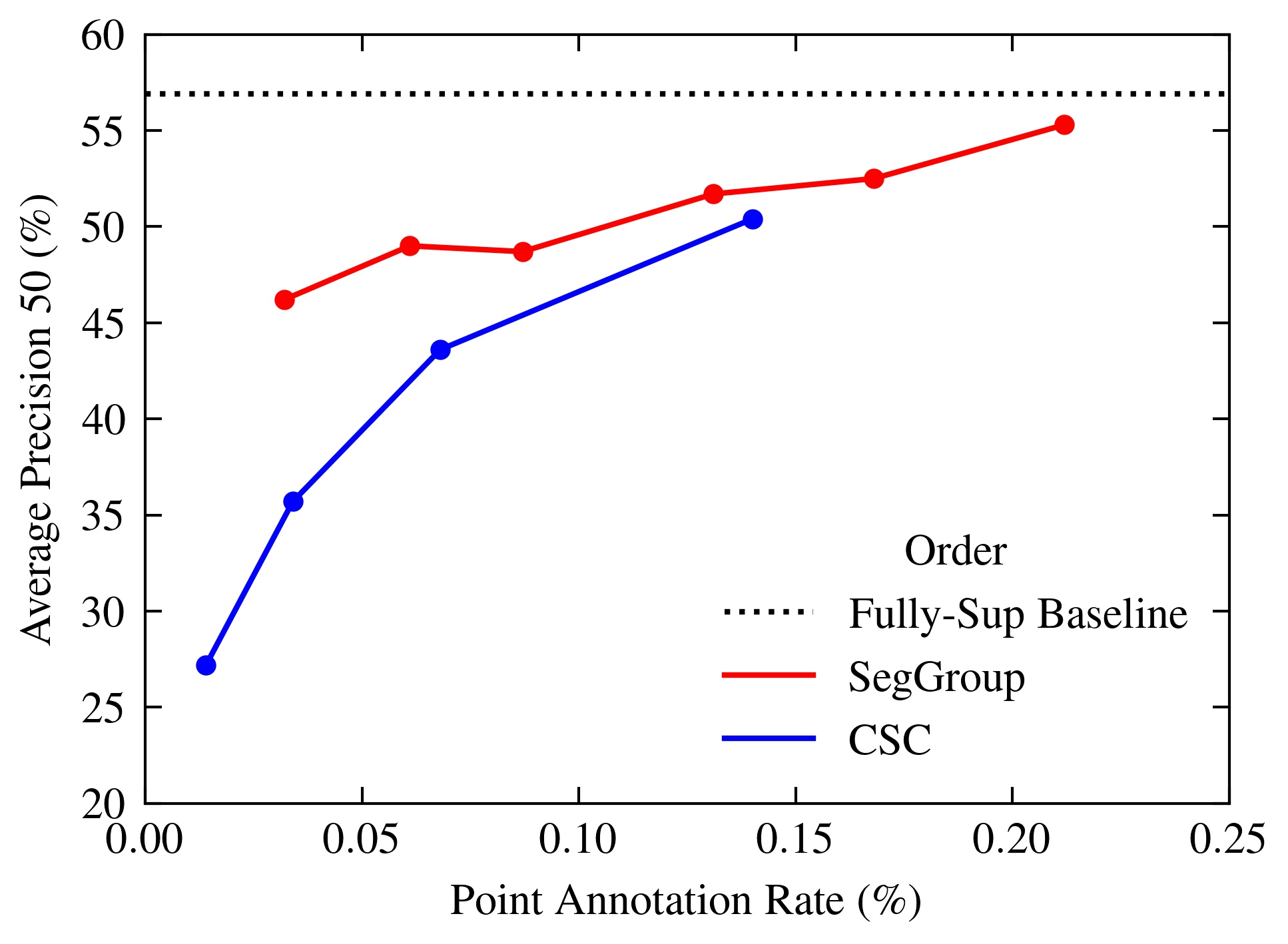}
	\end{center}
	\vspace{-5pt}
	\caption{Point cloud instance segmentation AP{\tiny 50} results (\%) of PointGroup~\cite{jiang2020pointgroup} network trained by SegGroup and CSC~\cite{hou2021exploring} with different point annotation rates on the ScanNet validation set.}
	\label{ap50}
\end{figure}

\begin{figure}[t]
	\begin{center}
		\includegraphics[width=0.8\linewidth]{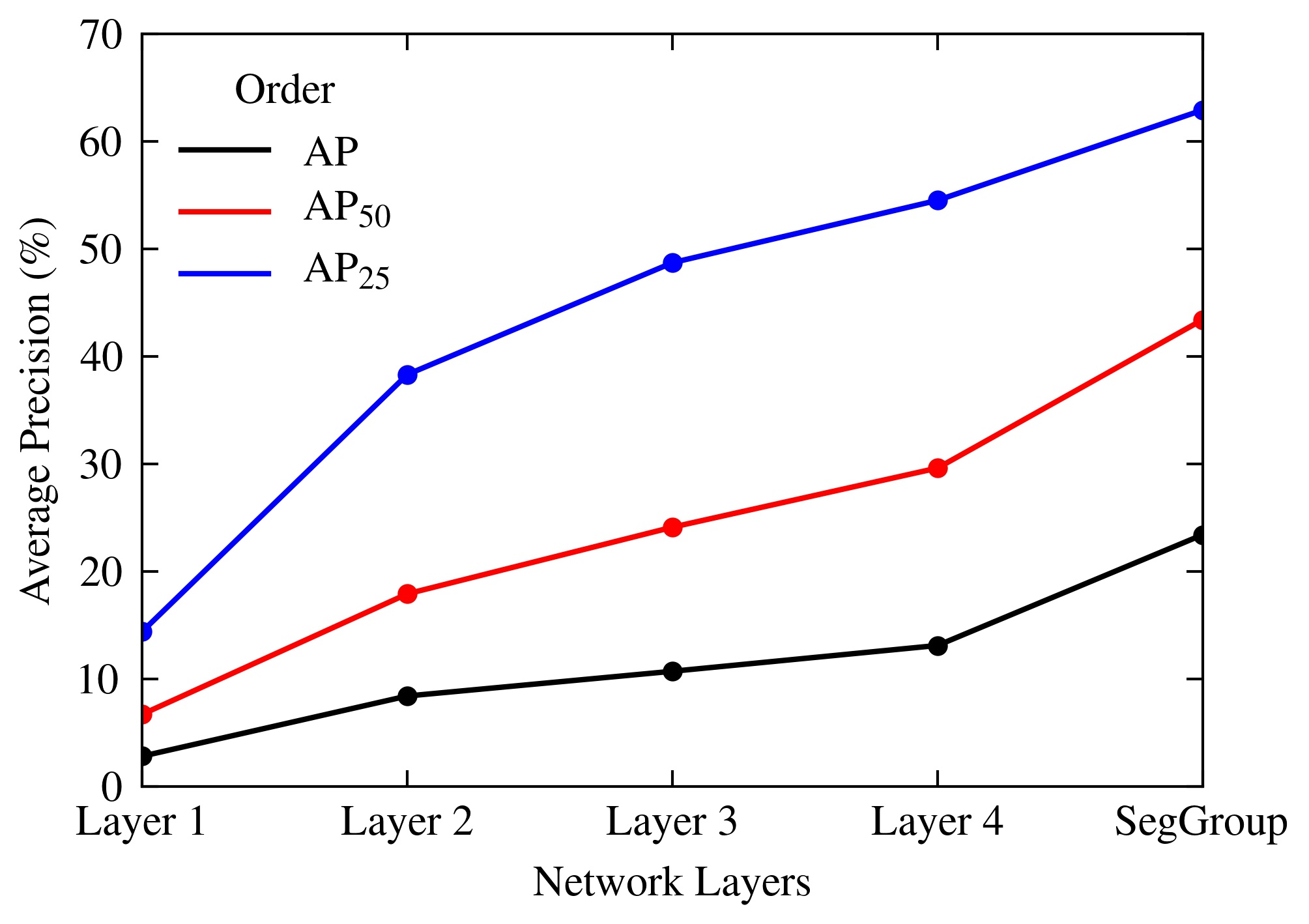}
	\end{center}
	\vspace{-10pt}
	\caption{Point cloud instance segmentation results (\%) of PointGroup network trained with pseudo labels from different layers of SegGroup on the ScanNet validation set. Layer names in X-axis are the same as the layer names in Table~\ref{sem_iou}.}
	\label{layer}
\end{figure}

When the annotation number increases, we find that we can achieve better performance in Table~\ref{anno_num}. We increased the annotation number by allowing more than one annotation for each instance. For fairness, we only compared results of mechanical seg-level labels generated from the largest $N$ segments for every instance. Following the settings in Table~\ref{label_type}, we also employed PointGroup~\cite{jiang2020pointgroup} and KPConv~\cite{thomas2019kpconv} to train a point cloud instance segmentation model and a point cloud semantic segmentation model.
%The experimental results are shown in Table~\ref{anno_num}. 
Both instance and semantic segmentation results show that the performance can be further boosted to be closer to the fully supervised baseline by the addition of annotation numbers per instance. When annotating ten segments per instance, we can even reach the same semantic segmentation performance with the fully supervised baseline. In Figure~\ref{ap50}, we also compare the AP{\tiny 50} results of our SegGroup in Table~\ref{anno_num} with the AP{\tiny 50} results of CSC~\cite{hou2021exploring} on instance segmentation. Experimental results validate the effectiveness of our method on the overlapping portion of the point annotation rate.

We also compared the point cloud instance segmentation performance using the generated pseudo labels from different layers of SegGroup. Fig.~\ref{layer} shows the results of PointGroup~\cite{jiang2020pointgroup} network trained with pseudo labels from different layers of SegGroup network on the ScanNet validation set. Layer names in Fig.~\ref{layer} are the same as the layer names in Table~\ref{sem_iou}. As the number of labeled segments increases during pseudo label generation in Table~\ref{sem_iou}, the segmentation performance in Fig.~\ref{layer} becomes better. 

%As human annotators may not accurately choose the largest segment for every instance, results are similar with the Top-3 segments selected by machines. Moreover, for the labels generated mechanically, the performance is similar for Top-1 and Top-3 labeling strategies. The performance gap between manual Top-1 and mechanical Top-3 annotations shows that semantic segmentation is more sensitive with annotation manner. Therefore, seg-level labels are practical for annotators to click on one point as the location on the most representative segment of an instance in 3D scene. 

%Because our annotation framework (the WebGL interface) can annotate more than one point for each instance, we further tested the model performance with different annotation numbers. Table~\ref{anno_num} shows the point cloud instance and semantic segmentation results under various annotation numbers per instance on the ScanNet validation set. For fairness, we only compare results of mechanical seg-level labels generated from the largest $N$ segments for every instance. Both instance and semantic segmentation results show that the performance can be boosted by the addition of annotation numbers per instance. 

\section{Conclusion}
In this paper, we have exploited the importance of instance locations for 3D scene segmentation and designed a weakly-supervised point cloud segmentation task for full exploitation. More specifically, we click on one point per instance to indicate its location and extend these location annotations into segments as seg-level labels by over-segmentation. We further design a segment grouping network (SegGroup) to generate point-level pseudo labels by grouping seg-level annotated segments into instances hierarchically, so that existing strong-supervised methods can directly consume the pseudo labels for training. Experimental results on both instance and semantic segmentation show that SegGroup effectively generates high-quality point-level pseudo labels from the locations of instances given the seg-level labels, which well balances the annotation cost and segmentation accuracy. %In future work, we plan to keep investigating the importance of instance locations in 3D scene related tasks. 

\ifCLASSOPTIONcaptionsoff
\newpage
\fi



%\bibliographystyle{IEEEtran}
%\begin{thebibliography}{1}
%\end{thebibliography}

\bibliographystyle{IEEEtran}
\bibliography{egbib.bib}


\vspace{30pt}
\begin{IEEEbiography}[{\includegraphics[width=1in,height=1.25in,clip,keepaspectratio]{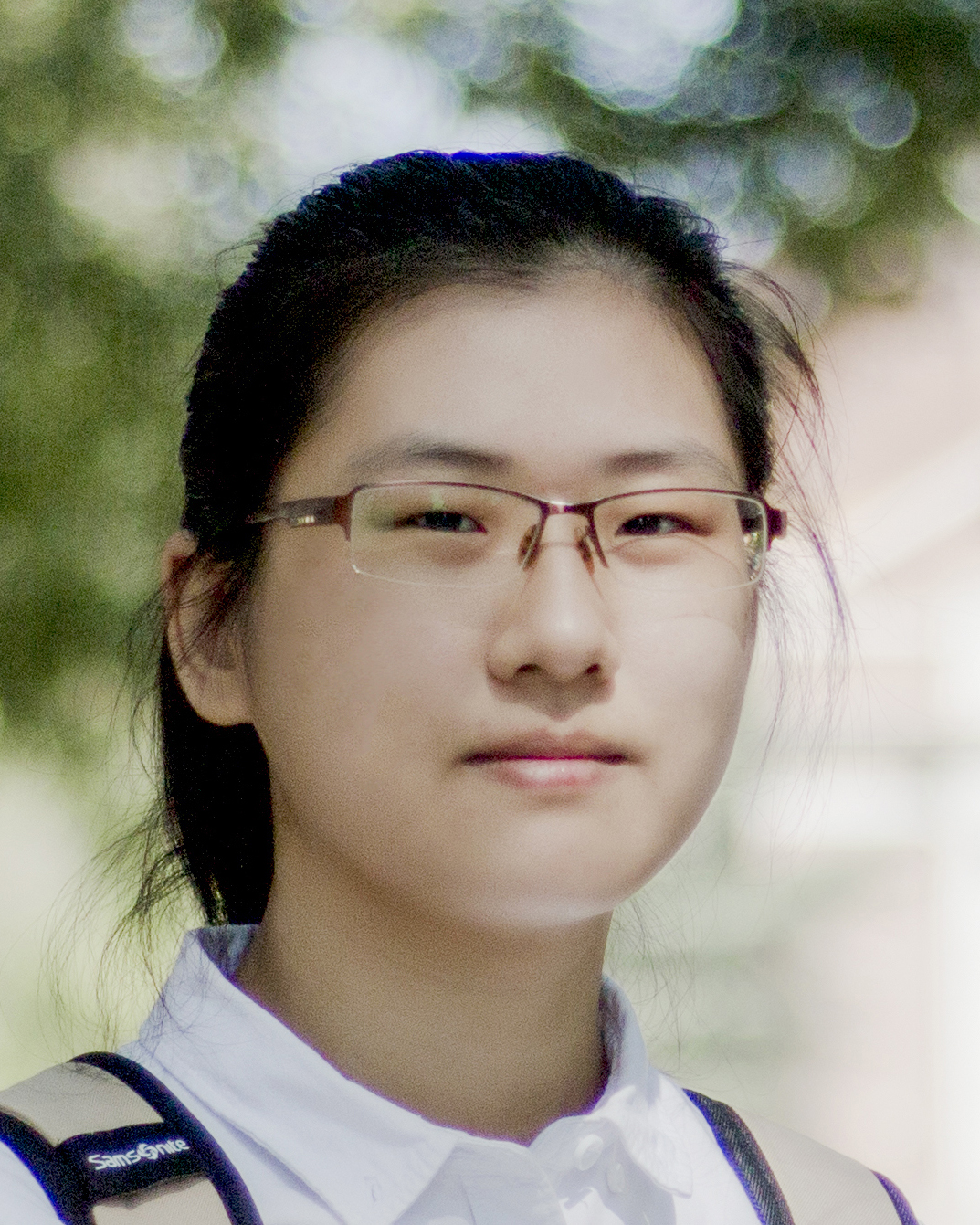}}]{An Tao}
	(Graduate Student Member, IEEE) received the B.Eng. degree from the School of Information Science and Engineering, Southeast University, Nanjing, China, in 2019. She is currently pursuing the Ph.D. degree in the Department of Automation, Tsinghua University, Beijing, China. Her current research interest is 3D vision. She has served as a reviewer for several conferences, e.g., CVPR, ICCV, ECCV, ICME, and 3DV.
\end{IEEEbiography}

\vspace{30pt}
\begin{IEEEbiography}[{\includegraphics[width=1in,height=1.25in,clip,keepaspectratio]{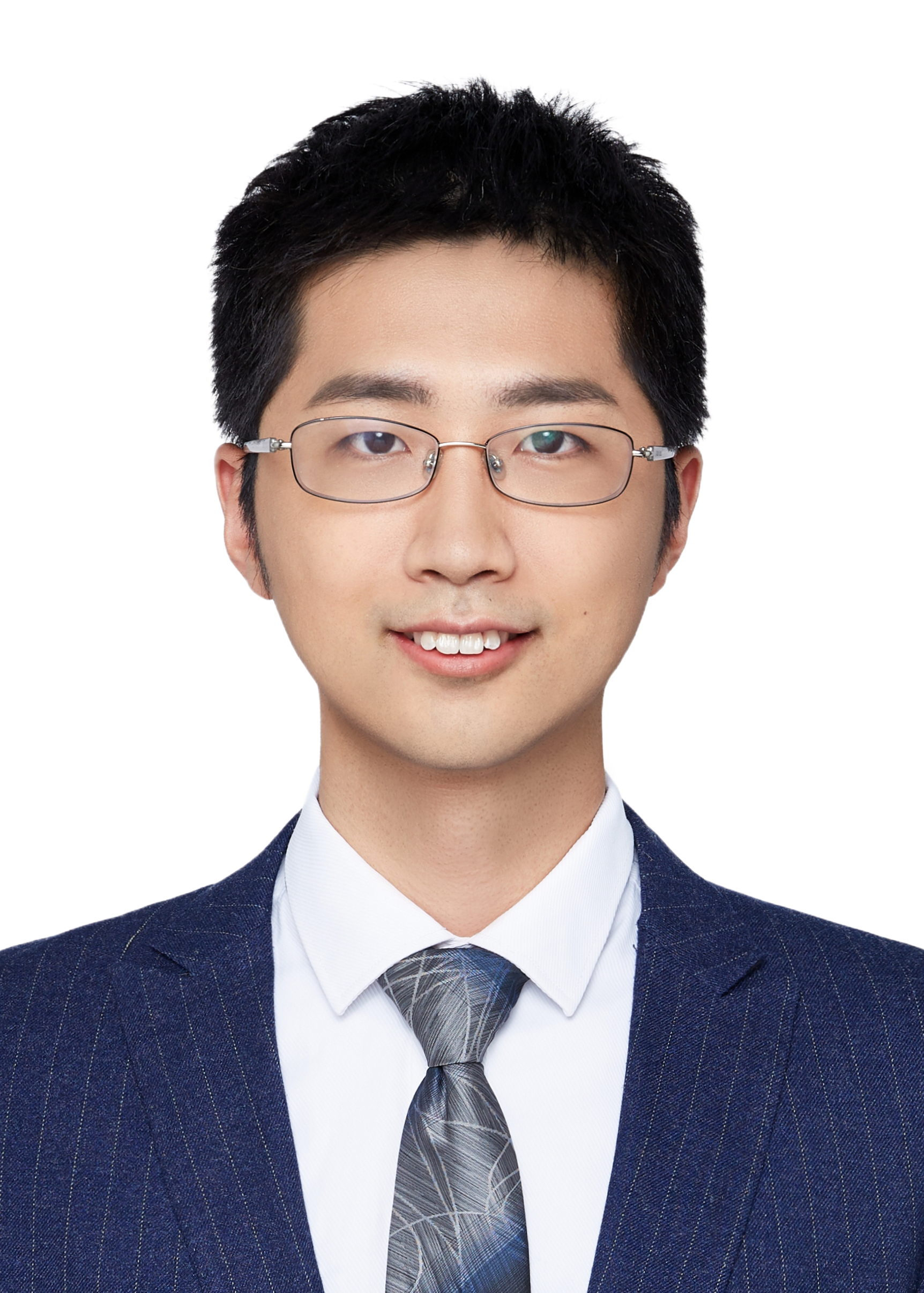}}]{Yueqi Duan}
	(Member, IEEE) received the B.S. and Ph.D. degrees from the Department of Automation, Tsinghua University, in 2014 and 2019, respectively. From 2019 to 2021, he served as a Postdoctoral Researcher with the Computer Science Department, Stanford University. He is currently an assistant professor with the Department of Electronic Engineering, Tsinghua University. His research interests include computer vision and pattern recognition. He has published 10+ scientific articles in the top journals and conferences including TPAMI, TIP, CVPR and ECCV. He serves as an area chair of ICME from 2020 to 2022, and a regular reviewer for a number of journals and conferences, e.g., TPAMI, IJCV, TIP, CVPR, ICCV, ECCV, ICML, NeurIPS and SIGGRAPH. He was awarded the Excellent Doctoral Dissertation of Chinese Association for Artificial Intelligence (CAAI) in 2020.
\end{IEEEbiography}

\vspace{30pt}
\begin{IEEEbiography}[{\includegraphics[width=1in,height=1.25in,clip,keepaspectratio]{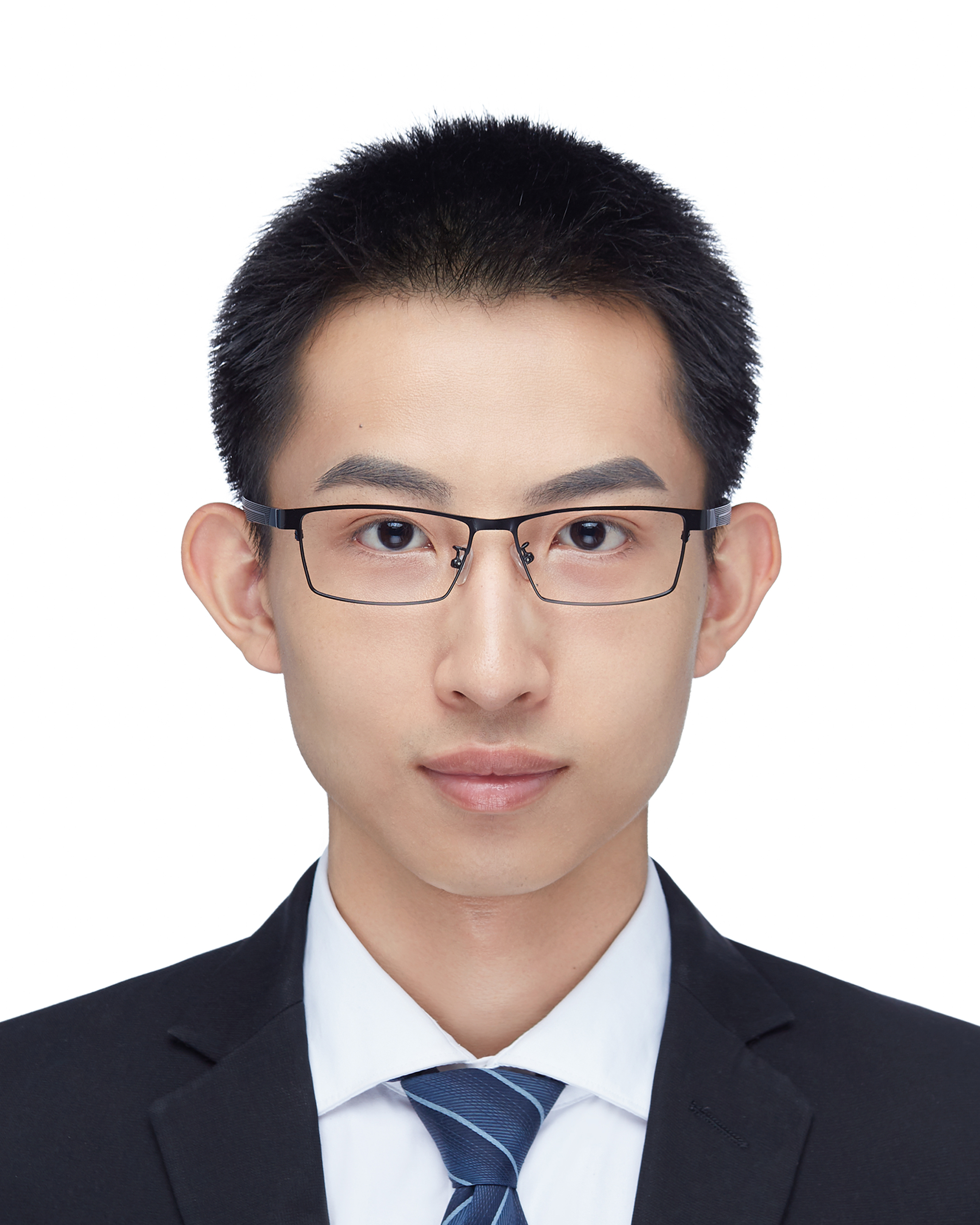}}]{Yi Wei}
	received the B.S. degree from the Department of Electronic Engineering, Tsinghua University, Beijing, China, in 2019, where he is currently pursuing the Ph.D. degree with the Department of Automation. His research interests lie in 3D vision, computer graphics, and robotics, especially focusing on 3D scene understanding and 3D reconstruction. He has authored six scientific articles in this area, including CVPR, ECCV, and ICRA. He serves as a reviewer for a number of journals and conferences, e.g., TIP, TCSVT, CVPR, and ICCV.
	
\end{IEEEbiography}\vfill

\begin{IEEEbiography}[{\includegraphics[width=1in,height=1.25in,clip,keepaspectratio]{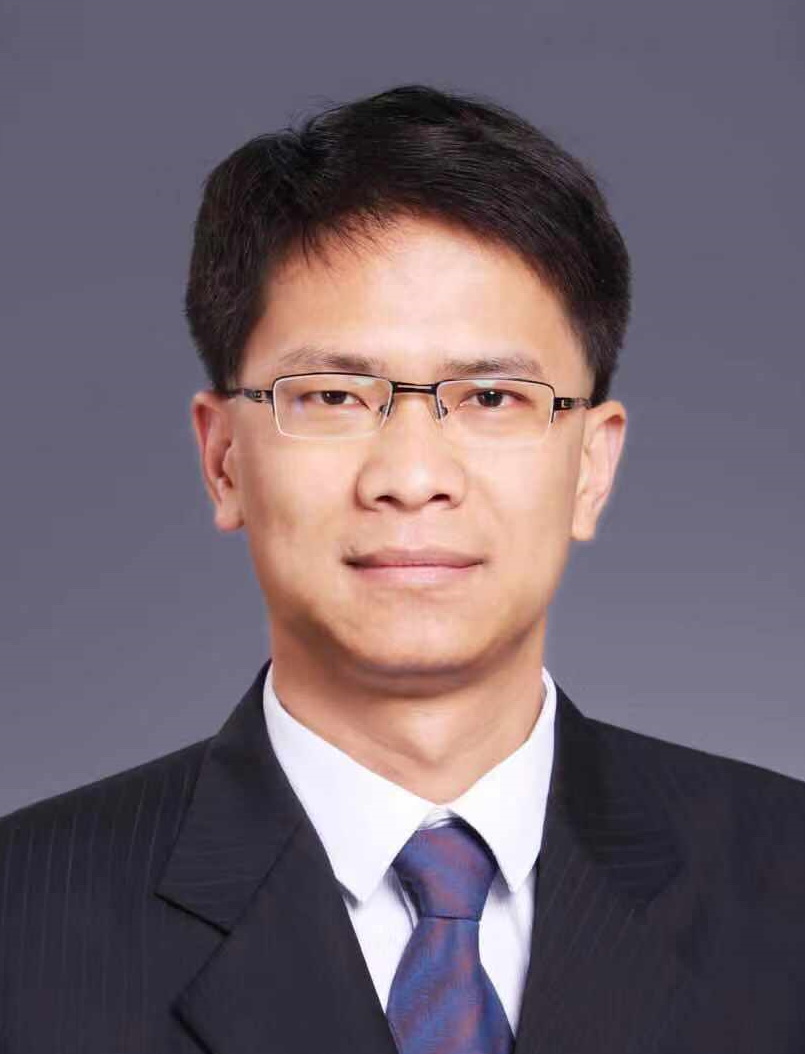}}]{Jiwen Lu}
	(Senior Member, IEEE) received the B.Eng. degree in mechanical engineering and the M.Eng. degree in electrical engineering from the Xi’an University of Technology, Xi’an, China, in 2003 and 2006, respectively, and the Ph.D. degree in electrical engineering from Nanyang Technological University, Singapore, in 2012. He is currently an Associate Professor with the Department of Automation, Tsinghua University, Beijing, China. His current research interests include computer vision and pattern recognition. He was/is a member of the Image, Video and Multidimensional Signal Processing Technical Committee, the Multimedia Signal Processing Technical Committee, the Information Forensics and Security Technical Committee of the IEEE Signal Processing Society, the Multimedia Systems and Applications Technical Committee, and the Visual Signal Processing and Communications Technical Committee of the IEEE Circuits and Systems Society. He is a fellow of IAPR. He was a recipient of the National Natural Science Funds for Distinguished Young Scholar. He serves as the General Co-Chair for the International Conference on Multimedia and Expo (ICME) 2022 and the Program Co-Chair for the ICME 2020, the International Conference on Automatic Face and Gesture Recognition (FG) 2023, and the International Conference on Visual Communication and Image Processing (VCIP) 2022. He serves as the Co-Editor-of-Chief for \textit{Pattern Recognition Letters} and an Associate Editor for the \textsc{IEEE Transactions on Image Processing}, the \textsc{IEEE Transactions on Circuits and Systems for Video Technology}, and the \textsc{IEEE Transactions on Biometrics, Behavior, and Identity Sciences}.
	
\end{IEEEbiography}

\vspace{30pt}
\begin{IEEEbiography}[{\includegraphics[width=1in,height=1.25in,trim=100 500 100 0,clip,keepaspectratio]{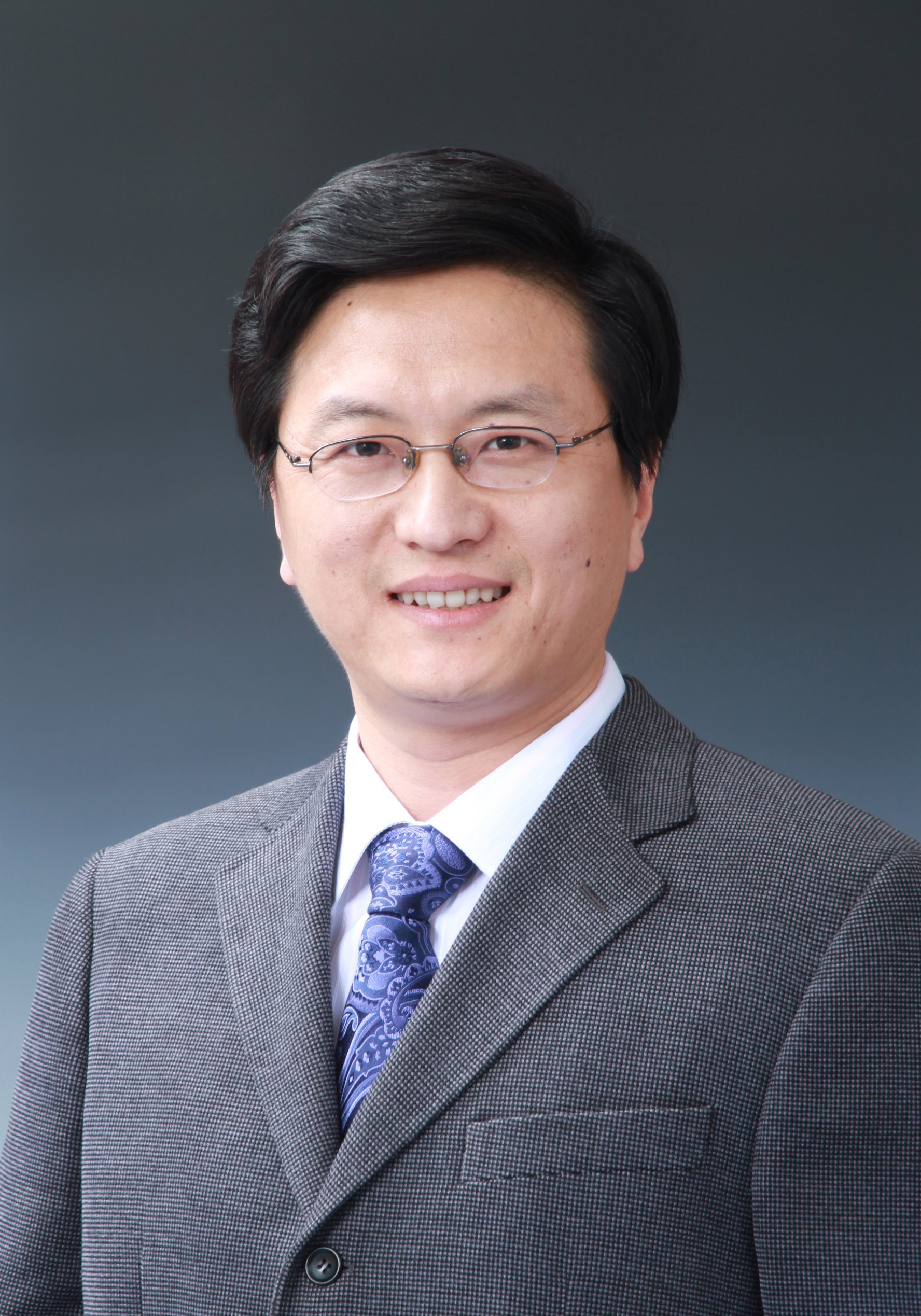}}]{Jie Zhou}
	(Senior Member, IEEE) received the B.S. and M.S. degrees from the Department of Mathematics, Nankai University, Tianjin, China, in 1990 and 1992, respectively, and the Ph.D. degree from the Institute of Pattern Recognition and Artificial Intelligence, Huazhong University of Science and Technology (HUST), Wuhan, China, in 1995. From 1995 to 1997, he worked as a Postdoctoral Fellow with the Department of Automation, Tsinghua University, Beijing, China. Since 2003, he has been a Full Professor with the Department of Automation, Tsinghua University. In recent years, he has authored more than 100 papers in peer-reviewed journals and conferences. Among them, more than 40 papers have been published in top journals and conferences, such as the \textsc{IEEE Transactions on Pattern Analysis and Machine Intelligence}, the \textsc{IEEE Transactions on Image Processing}, and CVPR. His research interests include computer vision, pattern recognition, and image processing. He is an IAPR Fellow. He received the National Outstanding Youth Foundation of China Award. He is an Associate Editor of the \textsc{IEEE Transactions on Pattern Analysis and Machine Intelligence} and two other journals.
\end{IEEEbiography}


\vfill


\end{document}